\documentclass[10pt,twocolumn,letterpaper]{article}

\usepackage{iccv}
\usepackage{times}
\usepackage{epsfig}
\usepackage{graphicx}
\usepackage{amsmath}
\usepackage{amssymb}

\usepackage[normalem]{ulem}
\usepackage{multirow,booktabs}
\usepackage{footnote}
\usepackage{kotex}
\usepackage{array}
\usepackage{adjustbox}
\usepackage[table]{xcolor}
\usepackage[flushleft]{threeparttable}
\usepackage{subcaption}

\definecolor{gold}{RGB}{212,175,55}

\usepackage[pagebackref=true,breaklinks=true,letterpaper=true,colorlinks,bookmarks=false]{hyperref}

\iccvfinalcopy 

\def\iccvPaperID{****} 

\begin{document}

\title{What Is Wrong With Scene Text Recognition Model Comparisons? \\
Dataset and Model Analysis}

\author{
  Jeonghun Baek\textsuperscript{1}\qquad\qquad
  Geewook Kim\textsuperscript{2}\thanks{Work performed as an intern in Clova AI Research.}\qquad\qquad
  Junyeop Lee\textsuperscript{1}\qquad\qquad
  Sungrae Park\textsuperscript{1}\\
  Dongyoon Han\textsuperscript{1}\qquad\qquad
  Sangdoo Yun\textsuperscript{1}\qquad\qquad
  Seong Joon Oh\textsuperscript{1}\qquad\qquad
  Hwalsuk Lee\textsuperscript{1}\thanks{Corresponding author.}\\
  \textsuperscript{1}Clova AI Research, NAVER/LINE Corp. \qquad\textsuperscript{2}Kyoto University\\
  {\tt\small \{jh.baek,junyeop.lee,sungrae.park,dongyoon.han,sangdoo.yun,hwalsuk.lee\}@navercorp.com}\\
  {\tt\small geewook@sys.i.kyoto-u.ac.jp}\quad
  {\tt\small coallaoh@linecorp.com}\\
}

\maketitle

\begin{abstract}
Many new proposals for scene text recognition (STR) models have been introduced in recent years. 
While each claim to have pushed the boundary of the technology, a holistic and fair comparison has been largely missing in the field due to the inconsistent choices of training and evaluation datasets.
This paper addresses this difficulty with three major contributions.
First, we examine the inconsistencies of training and evaluation datasets, and the performance gap results from inconsistencies.
Second, we introduce a unified four-stage STR framework that most existing STR models fit into. Using this framework allows for the extensive evaluation of previously proposed STR modules and the discovery of previously unexplored module combinations. 
Third, we analyze the module-wise contributions to performance in terms of accuracy, speed, and memory demand, under one consistent set of training and evaluation datasets. 
Such analyses clean up the hindrance on the current comparisons to understand the performance gain of the existing modules.
Our code is publicly available\footnote{\url{https://github.com/clovaai/deep-text-recognition-benchmark}}.

\end{abstract}

\vspace{-4mm}
\section{Introduction}\label{Introduction}
Reading text in natural scenes, referred to as scene text recognition (STR), has been an important task in a wide range of industrial applications.
The maturity of Optical Character Recognition (OCR) systems has led to its successful application on cleaned documents, but most traditional OCR methods have failed to be as effective on STR tasks due to the diverse text appearances that occur in the real world and the imperfect conditions in which these scenes are captured. 

To address these challenges, prior works~\cite{CRNN,RARE,lee2016recursive,liu2016star,GRCNN,yang2017learning,FAN,liu2018char,cheng2018aon,Bai_2018_CVPR,Rosetta,Liu_2018_ECCV} have proposed multi-stage pipelines, where each stage is a deep neural network addressing a specific challenge. 
For example, Shi \etal~\cite{CRNN} have suggested using a recurrent neural network to address the varying number of characters in a given input, and a connectionist temporal classification loss~\cite{CTC} to identify the number of characters.
Shi \etal~\cite{RARE} have proposed a transformation module that normalizes the input into a straight text image to reduce the representational burden for downstream modules to handle curved texts.

However, it is hard to assess whether and how a newly proposed module improves upon the current art, as some papers have come up with different evaluation and testing environments, making it difficult to compare reported numbers at face value (Table~\ref{tab:existing-work}).
We observed that 1) the training datasets and 2) the evaluation datasets deviate amongst various methods, as well.
For example, different works use a different subset of the IC13 dataset as part of their evaluation set, which may cause a performance disparity of more than 15\%.
This kind of discrepancy hinders the fair comparison of performance between different models.

\begin{table*}[h] 
  \tabcolsep=0.13cm
	\begin{center}
        \begin{adjustbox}{width=0.975\textwidth}
        \begin{tabular}{cl|c|l|cccccccccc|r|r}
			\hline
			& \multirow{2}{*}{Model} & \multirow{2}{*}{Year} 
			& \multirow{2}{*}{Train data} & IIIT & SVT & \multicolumn{2}{c}{IC03} & \multicolumn{2}{c}{IC13} & \multicolumn{2}{c}{IC15} & SP & CT & Time  & params \cr
			& &&& 3000 & 647 & 860 & 867 & 857 & 1015  & 1811 & 2077 & 645 & 288 & ms/image & $\times10^6$  \cr\hline
\parbox[t]{2mm}{\multirow{12}{*}{\rotatebox[origin=c]{90}{\textbf{Reported results}}}}
&CRNN \cite{CRNN}  & 2015    &  MJ  &78.2 & 80.8 & 89.4& $-$ & $-$ &  86.7 & $-$ & $-$ & $-$ & $-$ & 160 & 8.3\cr
&RARE  \cite{RARE} & 2016   & MJ  & 81.9 & 81.9 & 90.1 & $-$  & 88.6 & $-$ & $-$ & $-$ & 71.8 & 59.2 & $<$2  & $-$ \cr
&R2AM~\cite{lee2016recursive} & 2016 & MJ & 78.4 & 80.7 &  88.7 & $-$ & $-$ & 90.0 & $-$ & $-$ & $-$ & $-$ & 2.2 & $-$ \\
&STAR-Net~\cite{liu2016star} & 2016 &  MJ+PRI & 83.3 & 83.6 & 89.9 & $-$  & $-$ & 89.1  & $-$ & $-$ & 73.5 & $-$ & $-$ & $-$ \\
&GRCNN~\cite{GRCNN} &  2017  & MJ & 80.8 & 81.5 & 91.2 & $-$ & $-$& $-$& $-$ & $-$ & $-$ & $-$ & $-$ & $-$\\
& ATR \cite{yang2017learning} & 2017 &  PRI+C & $-$ & $-$ & $-$ & $-$ & $-$ & $-$ & $-$ & $-$ & \textbf{75.8} & 69.3 & $-$ & $-$ \\
& FAN  \cite{FAN} & 2017   & MJ+ST+C & 87.4 & 85.9 & $-$ & 94.2  & $-$ & 93.3  & 70.6 & $-$ & $-$  & $-$ & $-$ & $-$ \\
& Char-Net \cite{liu2018char} & 2018 & MJ & 83.6 & 84.4  & \textbf{91.5} & $-$ &  90.8 & $-$ & $-$ & 60.0 & 73.5 & $-$ & $-$ & $-$\\
& AON \cite{cheng2018aon} & 2018 & MJ+ST & 87.0 & 82.8 & $-$ & 91.5  & $-$ & $-$ & $-$ & \textbf{68.2} & 73.0 & \textbf{76.8} & $-$ & $-$\\
& EP \cite{Bai_2018_CVPR} & 2018 & MJ+ST & 88.3 & \textbf{87.5} & $-$ & 94.6 & $-$ & \textbf{94.4}  & \textbf{73.9} & $-$ & $-$ & $-$  & $-$ & $-$\\
&Rosetta \cite{Rosetta} & 2018 & PRI & $-$ & $-$ & $-$ & $-$ & $-$ & $-$ & $-$ & $-$ & $-$ & $-$ & $-$  & $-$ \\
& SSFL \cite{Liu_2018_ECCV} & 2018 & MJ & \textbf{89.4} & 87.1 & $-$ & \textbf{94.7} & \textbf{94.0} & $-$ & $-$  & $-$ & 73.9 & 62.5& $-$ & $-$\\
\hline \hline
\parbox[t]{2mm}{\multirow{7}{*}{\rotatebox[origin=c]{90}{\textbf{Our experiment}}}}
&CRNN~\cite{CRNN}                & 2015 & MJ+ST & 82.9 & 81.6 & 93.1 & 92.6 & 91.1 & 89.2 & 69.4 & 64.2 & 70.0 & 65.5 & 4.4 & 8.3\cr
&RARE~\cite{RARE}                & 2016 & MJ+ST      & 86.2 & 85.8 & 93.9 & 93.7 & 92.6 & 91.1 & 74.5 & 68.9 & 76.2 & 70.4 & 23.6 & 10.8 \cr
&R2AM~\cite{lee2016recursive} & 2016 & MJ+ST  &83.4 & 82.4 & 92.2 & 92.0 & 90.2 & 88.1 & 68.9 & 63.6 & 72.1 & 64.9& 24.1 & 2.9\\
&STAR-Net~\cite{liu2016star}     & 2016 & MJ+ST &87.0 & 86.9 & 94.4 & 94.0 & 92.8 & 91.5 & 76.1 & 70.3 & 77.5 & 71.7 & 10.9 & 48.7\\
&GRCNN~\cite{GRCNN}              & 2017 & MJ+ST       &84.2 & 83.7 & 93.5 & 93.0 & 90.9 & 88.8 & 71.4 & 65.8 & 73.6 & 68.1 & 10.7 & 4.6 \\
&Rosetta \cite{Rosetta}          & 2018 & MJ+ST & 84.3 & 84.7 & 93.4 & 92.9 & 90.9 & 89.0 & 71.2 & 66.0 & 73.8 & 69.2& 4.7 & 44.3\\
\cline{2-16} 
&Our best model                           &   & MJ+ST  & \textbf{87.9} & \textbf{87.5} & \textbf{94.9} & \textbf{94.4} & \textbf{93.6} & \textbf{92.3} &\textbf{77.6} & \textbf{71.8} & \textbf{79.2} & \textbf{74.0} & 27.6 & 49.6\\
\hline
        \end{tabular}
        \end{adjustbox}       
    \caption{Performance of existing STR models with their \textbf{inconsistent} training and evaluation settings. This inconsistency hinders the fair comparison among those methods. 
    We present the results reported by the original papers and also show our re-implemented results under unified and consistent setting. 
    At the last row, we also show the best model we have found, which shows competitive performance to state-of-the-art methods.
    MJ, ST, C, and, PRI denote MJSynth~\cite{Jaderberg14c}, SynthText~\cite{Gupta16}, Character-labeled~\cite{FAN,yang2017learning}, and private data~\cite{Rosetta}, respectively.
    Top accuracy for each benchmark is shown in \textbf{bold}.
    }
    \label{tab:existing-work}
    \end{center}
    \vspace{-7mm}
\end{table*} 

Our paper addresses these types of issues with the following main contributions.
First, we analyze all training and evaluation datasets commonly used in STR papers. Our analysis reveals the inconsistency of using the STR datasets and its causes. 
For instance, we found 7 missing examples in IC03 dataset and 158 missing examples in IC13 dataset as well.
We investigate several previous works on the STR datasets and show that the inconsistency causes incomparable results as shown in Table~\ref{tab:existing-work}.
Second, we introduce a unifying framework for STR that provides a common perspective for existing methods. 
Specifically, we divide the STR model into four different consecutive stages of operations: transformation (Trans.), feature extraction (Feat.), sequence modeling (Seq.), and prediction (Pred.).
The framework provides not only existing methods but their possible variants toward an extensive analysis of module-wise contribution.
Finally, we study the module-wise contributions in terms of accuracy, speed, and memory demand, under a unified experimental setting. With this study, we assess the contribution of individual modules more rigorously and propose previously overlooked module combinations that improves over the state of the art.
Furthermore, we analyzed failure cases on the benchmark dataset to identify remaining challenges in STR.
\section{Dataset Matters in STR}\label{dataset}
In this section, we examine the different training and evaluation datasets used by prior works, and then their discrepancies are addressed.
Through this analysis, we highlight how each of the works differs in constructing and using their datasets, and investigate the bias caused by the inconsistency when comparing performance between different works (Table~\ref{tab:existing-work}).
The performance gaps due to dataset inconsistencies are measured through experiments and discussed in \S\ref{sec:experiments}.

\subsection{Synthetic datasets for training} \label{synthetic_data}
\begin{figure}
  \begin{subfigure}{0.235\textwidth} \centering
  	 \includegraphics[width=1.0\textwidth]{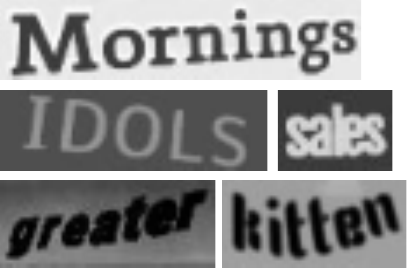}
     \caption{MJSynth word boxes}\label{MJSynth}
   \end{subfigure}
   \begin{subfigure}{0.235\textwidth} \centering
     \includegraphics[width=0.9\textwidth]{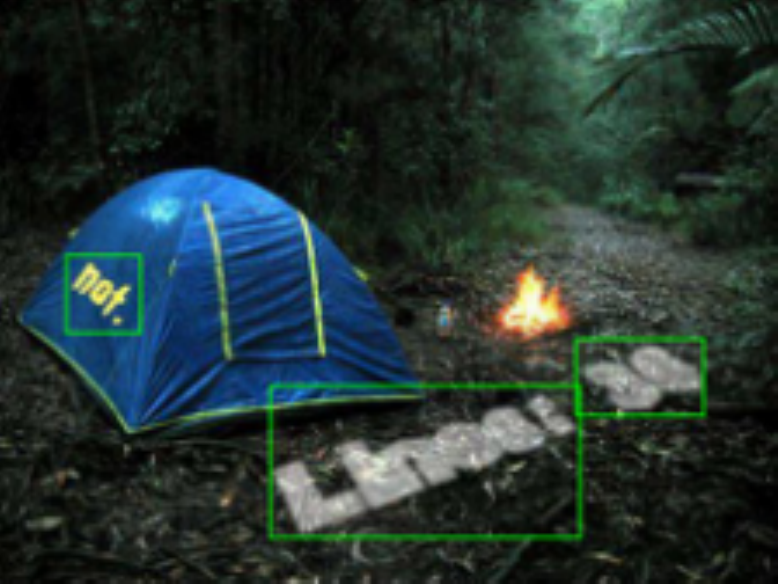}
     \caption{SynthText scene image}\label{SynthText}
   \end{subfigure}
  \caption{Samples of MJSynth and SynthText used as training data.}
  \label{Synthetic}
\end{figure}

When training a STR model, labeling scene text images is costly, and thus it is difficult to obtain enough labeled data for. Alternatively using real data, most STR models have used synthetic datasets for training. We first introduce two most popular synthetic datasets used in recent STR papers:
\begin{itemize}
    \item \noindent\textbf{MJSynth~(MJ)}~\cite{Jaderberg14c}
    is a synthetic dataset designed for STR, containing 8.9~M word box images.
    The word box generation process is as follows: 1) font rendering, 2) border and shadow rendering, 3) background coloring, 4) composition of font, border, and background, 5) applying projective distortions, 6) blending with real-world images, and 7) adding noise.
    Figure~\ref{MJSynth} shows some examples of MJSynth,
    \item \noindent\textbf{SynthText~(ST)}~\cite{Gupta16}
    is another synthetically generated dataset and was originally designed for scene text detection. An example of how the words are rendered onto scene images is shown in Figure~\ref{SynthText}. Even though SynthText was designed for scene text detection task, it has been also used for STR by cropping word boxes. SynthText has 5.5~M training data once the word boxes are cropped and filtered for non-alphanumeric characters.
\end{itemize}

Note that prior works have used diverse combinations of MJ, ST, and or other sources (Table~\ref{tab:existing-work}). These \textbf{inconsistencies} call into question whether the improvements are due to the contribution of the proposed module or to that of a better or larger training data. Our experiment in \S\ref{subsec:traindata} describes the influence of the training datasets to the final performance on the benchmarks. We further suggest that future STR researches clearly indicate the training datasets used and compare models using the same training set.

\subsection{Real-world datasets for evaluation}\label{benchmarks_datasets}
Seven real-world STR datasets have been widely used for evaluating a trained STR model. 
For some benchmark dataset, \textbf{different subsets} of the dataset may have been used in each prior work for evaluation (Table~\ref{tab:existing-work}). 
These difference in subsets result in \textbf{inconsistent} comparison. 

We introduce the datasets by categorizing them into regular and irreguglar datsets. 
The benchmark datasets are given the distinction of being ``regular'' or ``irregular'' datasets~\cite{RARE,yang2017learning,cheng2018aon}, according to the difficulty and geometric layout of the texts.
First, \textbf{\bf regular datasets} contain text images with horizontally laid out characters that have even spacings between them.
These represent relatively easy cases for STR:

\begin{itemize}
    \item \noindent\textbf{IIIT5K-Words}~\textbf{(IIIT)}~\cite{IIIT5K} is the dataset crawled from Google image searches, with query words that are likely to return text images, such as ``billboards'', ``signboard'', ``house numbers'', ``house name plates'', and ``movie posters''. IIIT consists of 2,000 images for training and 3,000 images for evaluation,
    \vspace{-2mm}
    \item \noindent\textbf{Street View Text}~\textbf{(SVT)}~\cite{SVT} contains outdoor street images collected from Google Street View. Some of these images are noisy, blurry, or of low-resolution. SVT consists of 257 images for training and 647 images for evaluation,
    \vspace{-2mm}
    \item \noindent\textbf{ICDAR2003}~\textbf{(IC03)}~\cite{IC03} was created for the ICDAR 2003 Robust Reading competition for reading camera-captured scene texts. It contains 1,156 images for training and 1,110 images for evaluation.
    Ignoring all words that are either too short (less than 3 characters) or ones that contain non-alphanumeric characters reduces 1,110 images to 867.
    However, researchers have used two different versions of the dataset for evaluation: versions with 860 and 867 images.
    The 860-image dataset is missing 7 word boxes compared to the 867 dataset.
    The omitted word boxes can be found in the supplementary materials,
    \vspace{-2mm}
    \item \noindent\textbf{ICDAR2013}~\textbf{(IC13)}~\cite{IC13} inherits most of IC03's images and was also created for the ICDAR 2013 Robust Reading competition.
    It contains 848 images for training and 1,095 images for evaluation, where pruning words with non-alphanumeric characters results in 1,015 images.
    Again, researchers have used two different versions for evaluation: 857 and 1,015 images.
    The 857-image set is a subset of the 1,015 set where words shorter than 3 characters are pruned.
\end{itemize}

Second, \textbf{\bf irregular datasets} typically contain harder corner cases for STR, such as curved and arbitrarily rotated or distorted texts~\cite{RARE,yang2017learning,cheng2018aon}:

\begin{figure}[t]
\centering
    \begin{subfigure}{0.49\linewidth} \centering
     \includegraphics[scale=0.6]{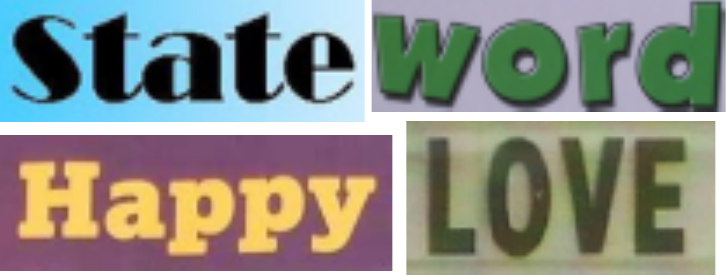}
     \caption{Regular}\label{fig:figA}
    \end{subfigure}
    \begin{subfigure}{0.49\linewidth} \centering
     \includegraphics[scale=0.6]{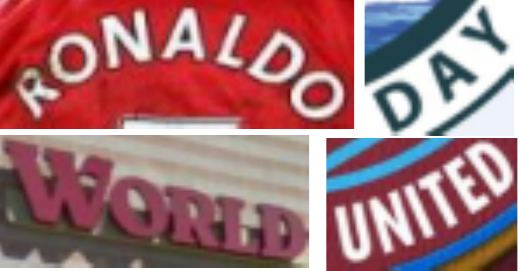}
     \caption{Irregular}\label{fig:figB}
    \end{subfigure}
\caption{Examples of regular (IIIT5k, SVT, IC03, IC13) and irregular (IC15, SVTP, CUTE) real-world datasets.} \label{fig:twofigs}
\end{figure}

\begin{figure*}[t]
\includegraphics[width=0.975\textwidth]{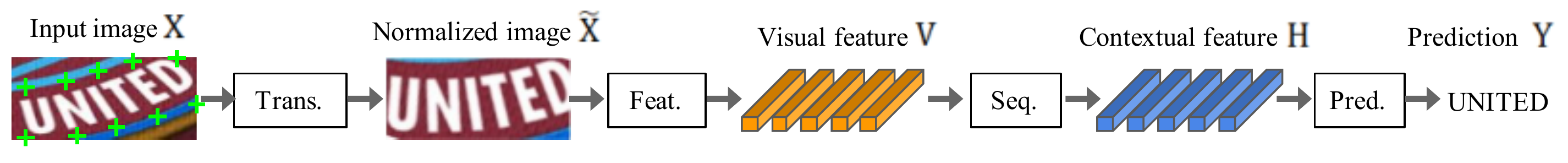}
	\centering
  \caption{Visualization of an example flow of scene text recognition. We decompose a model into four stages.}
  \label{4stage_flow}
\end{figure*}

\begin{itemize}
    \item \noindent\textbf{ICDAR2015}~\textbf{(IC15)}~\cite{IC15} was created for the ICDAR 2015 Robust Reading competitions and contains 4,468 images for training and 2,077 images for evaluation.
    The images are captured by Google Glasses while under the natural movements of the wearer. Thus, many are noisy, blurry, and rotated, and some are also of low resolution.
    Again, researchers have used two different versions for evaluation: 1,811 and 2,077 images.
    Previous papers~\cite{FAN,Bai_2018_CVPR} have only used 1,811 images, discarding non-alphanumeric character images and some extremely rotated, perspective-shifted, and curved images for evaluation. 
    Some of the discarded word boxes can be found in the supplementary materials,
    \item \noindent\textbf{SVT Perspective}~\textbf{(SP)}~\cite{SVTP} is collected from Google Street View and contains 645 images for evaluation.
    Many of the images contain perspective projections
    due to the prevalence of non-frontal viewpoints,
    \vspace{-2mm}
    \item \noindent\textbf{CUTE80}~\textbf{(CT)}~\cite{CUTE80} is collected from natural scenes and contains 288 cropped images for evaluation.
    Many of these are curved text images.
\end{itemize}

Notice that, Table~\ref{tab:existing-work} provides us a critical issue that
prior works evaluated their models on \textbf{different benchmark datasets}. Specifically,  the evaluation has been conducted on different versions of benchmarks in IC03, IC13 and IC15. In IC03, 7 examples can cause a performance gap by 0.8\% that is a huge gap when comparing those of prior performances. In the case of IC13 and IC15, the gap of the example numbers is even bigger than those of IC03.
\section{STR Framework Analysis}\label{model} 
The goal of the section is introducing the scene text recognition (STR) framework consisting of four stages, derived from commonalities among independently proposed STR models. 
After that, we describe the module options in each stage.

Due to the resemblance of STR to computer vision tasks (\eg object detection) and sequence prediction tasks, STR has benefited from high-performance convolutional neural networks (CNNs) and recurrent neural networks (RNNs). The first combined application of CNN and RNN for STR, Convolutional-Recurrent Neural Network (CRNN)~\cite{CRNN}, extracts CNN features from the input text image, and re-configures them with an RNN for robust sequence prediction. After CRNN, multiple variants \cite{RARE,lee2016recursive,liu2016star,liu2018char,GRCNN,FAN,Rosetta} have been proposed to improve performance. For rectifying arbitrary text geometries, as an example, transformation modules have been proposed to normalize text images~\cite{RARE,liu2016star,liu2018char}. For treating complex text images with high intrinsic dimensionality and latent factors (\eg font style and cluttered background), improved CNN feature extractors have been incorporated~\cite{lee2016recursive,GRCNN,FAN}. 
Also, as people have become more concerned with inference time, some methods have even omitted the RNN stage~\cite{Rosetta}. 
For improving character sequence prediction, attention based decoders have been proposed~\cite{lee2016recursive, RARE}.

The four stages derived from existing STR models are as follows:

\begin{enumerate}
\item \textbf{Transformation (Trans.)} normalizes the input text image using the Spatial Transformer Network~(STN~\cite{STN}) to ease downstream stages.
\vspace{-3mm}
\item \textbf{Feature extraction (Feat.)} maps the input image to a representation that focuses on the attributes relevant for character recognition, while suppressing irrelevant features such as font, color, size, and background. 
\vspace{-2mm}
\item \textbf{Sequence modeling (Seq.)} captures the contextual information within a sequence of characters for the next stage to predict each character more robustly, rather than doing it independently.
\vspace{-2mm}
\item \textbf{Prediction (Pred.)} estimates the output character sequence from the identified features of an image. 
\end{enumerate}
We provide Figure~\ref{4stage_flow} for an overview and all the architectures we used in this paper are found in the supplementary materials.

\noindent

\subsection{Transformation stage}\label{subsec:transformation}
The module of this stage transforms the input image $X$ into the normalized image $\tilde{X}$.
Text images in natural scenes come in diverse shapes, as shown by curved and tilted texts.
If such input images are fed unaltered, the subsequent feature extraction stage needs to learn an invariant representation with respect to such geometry.
To reduce this burden, thin-plate spline \textbf{(TPS)} transformation, a variant of the spatial transformation network (STN)~\cite{STN}, has been applied with its flexibility to diverse aspect ratios of text lines~\cite{RARE,liu2016star}. 
TPS employs a smooth spline interpolation between a set of \emph{fiducial points}.
More precisely, TPS finds multiple fiducial points (green '+' marks in Figure~\ref{4stage_flow}) at the upper and bottom enveloping points, and normalizes the character region to a predefined rectangle. 
Our framework allows for the selection or de-selection of TPS. 

\subsection{Feature extraction stage}
In this stage, a CNN abstract an input image (i.e., $X$ or $\tilde{X}$) and outputs a visual feature map $V = \{v_i\}, i=1,\dots,I$ ($I$ is the number of columns in the feature map).
Each column in the resulting feature map by a feature extractor has a corresponding distinguishable receptive field along the horizontal line of the input image. 
These features are used to estimate the character on each receptive field.

We study three architectures of \textbf{VGG}~\cite{VGG}, \textbf{RCNN}~\cite{lee2016recursive}, and \textbf{ResNet}~\cite{ResNet}, previously used as feature extractors for STR. 
VGG in its original form consists of multiple convolutional layers followed by a few fully connected layers~\cite{VGG}. 
RCNN is a variant of CNN that can be applied recursively to adjust its receptive fields depending on the character shapes~\cite{lee2016recursive,GRCNN}. 
ResNet is a CNN with \textit{residual connections} that eases the training of relatively deeper CNNs. 

\subsection{Sequence modeling stage}
The extracted features from Feat. stage are reshaped to be a sequence of features $V$.
That is, each column in a feature map $v_i\in V$ is used as a frame of the sequence.
However, this sequence may suffer the lack of contextual information.
Therefore, some previous works use Bidirectional LSTM \textbf{(BiLSTM)} to make a better sequence $H = \rm{Seq.}(V)$ after the feature extraction stage~\cite{CRNN,RARE,FAN}.
On the other hand, Rosetta~\cite{Rosetta} removed the BiLSTM to reduce computational complexity and memory consumption. 
Our framework allows for the selection or de-selection of BiLSTM. 

\subsection{Prediction stage}
In this stage, from the input $H$, a module predict a sequence of characters, (i.e., $Y=y_1,y_2,\dots$).
By summing up previous works, we have two options for prediction: 
(1) Connectionist temporal classification \textbf{(CTC)}~\cite{CTC}
and (2) attention-based sequence prediction \textbf{(Attn)}~\cite{RARE,FAN}. 
CTC allows for the prediction of a non-fixed number of a sequence even though a fixed number of the features are given. 
The key methods for CTC are to predict a character at each column ($h_i\in H$) and to modify the full character sequence into a non-fixed stream of characters by deleting repeated characters and \textit{blanks}~\cite{CTC,CRNN}. 
On the other hand, Attn automatically captures the information flow within the input sequence to predict the output sequence~\cite{NMT}. It enables an STR model to learn a character-level language model representing output class dependencies.

\begin{figure*}
   \centering
   \includegraphics[width=0.98\textwidth]{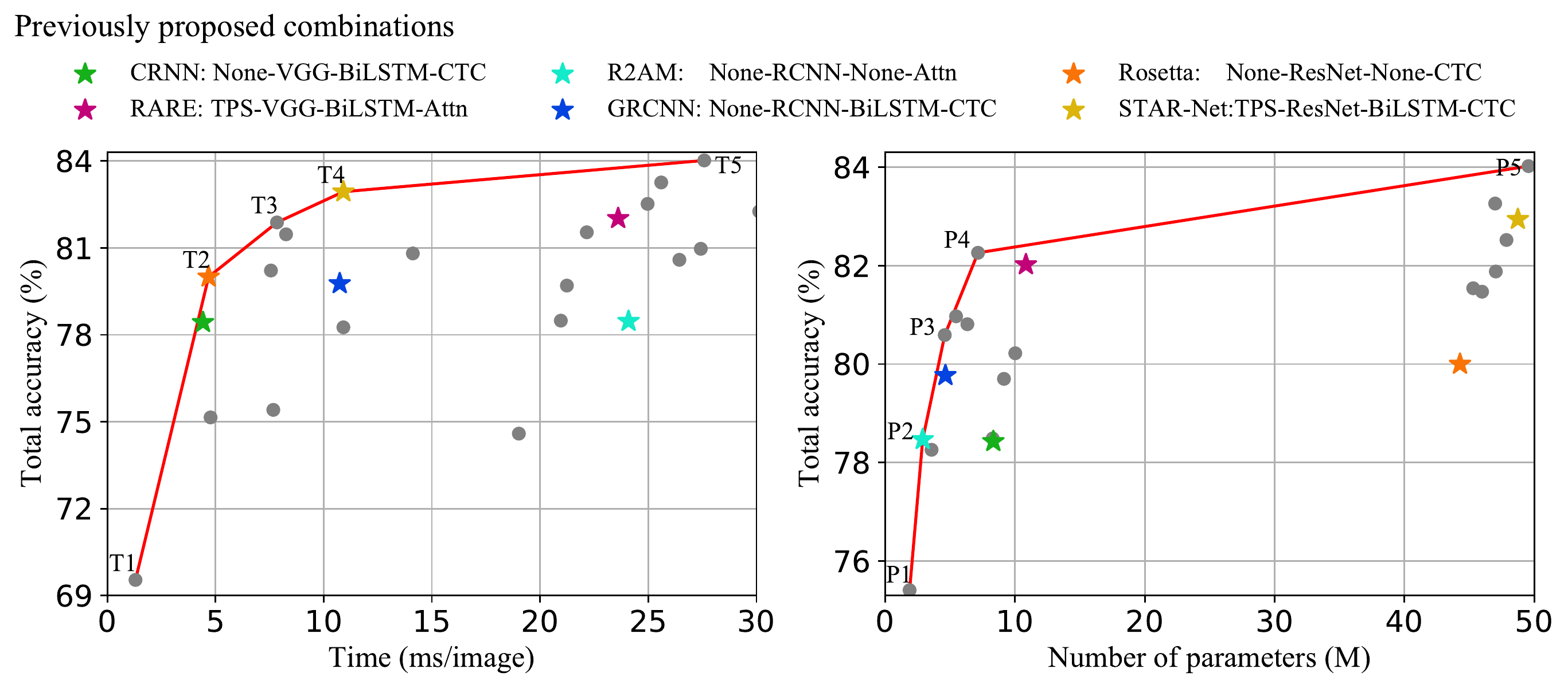}
    \begin{subfigure}{0.97\columnwidth}
    \scriptsize
    \begin{adjustbox}{width=1\textwidth}
	    \begin{tabular}{c|l|l|l|l|r|r|r}
			\hline
            \multirow{2}{*}{\#} & \multirow{2}{*}{Trans.} & \multirow{2}{*}{Feat.} & \multirow{2}{*}{Seq.} & \multirow{2}{*}{Pred.} &Acc. & Time & params \cr
  			 &&&&  & \% & ms  & $\times10^6$ \cr\hline
T1 & None & VGG & None & CTC &  69.5 & 1.3 & 5.6\\
T2 & None & \textbf{ResNet} & None & CTC & 80.0 & 4.7 & 46.0\\ 
T3 & None & ResNet & \textbf{BiLSTM} & CTC & 81.9 & 7.8 & 48.7\\ 
T4 & \textbf{TPS} & ResNet & BiLSTM & CTC & 82.9 & 10.9 & 49.6\\ 
T5 & TPS & ResNet & BiLSTM & \textbf{Attn} & 84.0 & 27.6 & 49.6\\ 
\hline
		\end{tabular}
		\end{adjustbox}
        \caption{Accuracy versus time trade-off curve and its frontier combinations}
    \label{tab:trade-off-time}
    \end{subfigure}
\hspace{3em}
\begin{subfigure}{0.97\columnwidth}
\scriptsize
    \begin{adjustbox}{width=1\textwidth}
	    \begin{tabular}{c|l|l|l|l|r|r|r}
			\hline
            \multirow{2}{*}{\#} & \multirow{2}{*}{Trans.} & \multirow{2}{*}{Feat.} & \multirow{2}{*}{Seq.} & \multirow{2}{*}{Pred.} & Acc. & Time & params \cr
  			 &&&&& \% & ms & $\times10^6$ \cr\hline
P1 & None & RCNN & None & CTC& 75.4 & 7.7 & 1.9\\
P2 & None & RCNN & None & \textbf{Attn}  & 78.5 & 24.1 & 2.9\\
P3 & \textbf{TPS} & RCNN & None & Attn  & 80.6 & 26.4 & 4.6\\
P4 & TPS & RCNN & \textbf{BiLSTM} & Attn  & 82.3 & 30.1 & 7.2\\
P5 & TPS & \textbf{ResNet} & BiLSTM & Attn  & 84.0 & 27.6 & 49.6\\
\hline
		\end{tabular}
	    \end{adjustbox}
        \caption{Accuracy versus memory trade-off curve and its frontier combinations}
    \label{tab:trade-off-memory}
    \end{subfigure}
    \caption{
   Two types of trade-offs exhibited by STR module combinations. 
   Stars indicate previously proposed models and circular dots represent new module combinations evaluated by our framework.
   Red solid curves indicate the trade-off frontiers found among the combinations.
   Tables under each plot describe module combinations and their performance on the trade-off frontiers. Modules in bold denote those that have been changed from the combination directly before it; those modules improve performance over the previous combination while minimizing the added time or memory cost.
   }
  \label{tradeoff}
\end{figure*}
\section{Experiment and Analysis} \label{sec:experiments}
This section contains the evaluation and analysis of all possible STR module combinations (2$\times$3$\times$2$\times$2$=$ 24 in total) from the four-stage framework in \S\ref{model}, all evaluated under the common training and evaluation dataset constructed from the datasets listed in \S\ref{dataset}.

\subsection{Implementation detail}\label{implementation}
As we described in \S\ref{dataset}, training and evaluation datasets influences the measured performances of STR models significantly. 
To conduct a fair comparison, we have fixed the choice of training, validation, and evaluation datasets.

\noindent\textbf{STR training and model selection}
We use an union of MJSynth 8.9~M and SynthText 5.5~M (14.4~M in total) as our training data.
We adopt the AdaDelta~\cite{zeiler2012adadelta} optimizer, whose decay rate is set to $\rho=0.95$. 
The training batch size is $192$, and the number of iterations is 300~K. Gradient clipping is used at magnitude 5. All parameters are initialized with He's method~\cite{he2015delving}.
We use the union of the training sets IC13, IC15, IIIT, and SVT as the validation data, and validated the model after every 2000 training steps to select the model with the highest accuracy on this set. 
Notice that, the validation set does not contain the IC03 train data because some of them were duplicated in the evaluation dataset of IC13. 
The total number of duplicated scene images is 34, and they contain 215 word boxes.
Duplicated examples can be found in the supplementary materials.

\noindent\textbf{Evaluation metrics}
In this paper, we provide a thorough analysis on STR combinations in terms of accuracy, time, and memory aspects altogether. For accuracy, we measure the success rate of word predictions per image on the 9 real-world evaluation datasets involving all subsets of the benchmarks, as well as a unified evaluation dataset (8,539 images in total); 3,000 from IIIT, 647 from SVT, 867 from IC03, 1015 from IC13, 2,077 from IC15, 645 from SP, and 288 from CT.
We only evaluate on alphabets and digits.
For each STR combination, we have run five trials with different initialization random seeds and have averaged their accuracies.
For speed assessment, we measure the per-image average clock time (in millisecond) for recognizing the given texts under the same compute environment, detailed below. 
For memory assessment, we count the number of trainable floating point parameters in the entire STR pipeline.

\noindent\textbf{Environment}:\;\:
For a fair speed comparison, all of our evaluations are performed on the same environment: an Intel Xeon(R) E5-2630 v4 2.20GHz CPU, an NVIDIA TESLA P40 GPU, and 252GB of RAM.
All experiments are performed with NAVER Smart Machine Learning (NSML) platform~\cite{kim2018nsml}.

\begin{figure}
 \includegraphics[width=0.475\textwidth]{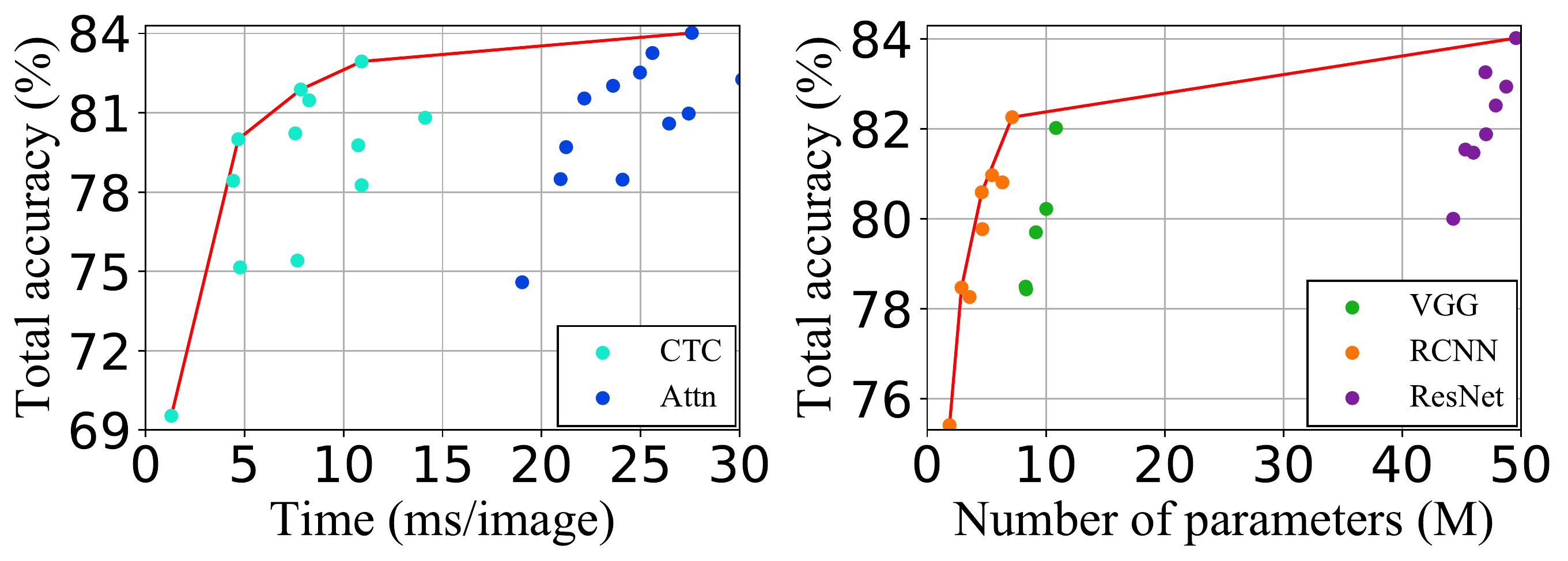}
 \caption{Color-coded version of Figure~\ref{tradeoff}, according to the prediction (left) and feature extraction (right) modules. They are identified as the most significant factors for speed and memory, respectively. 
 }
 \label{acc_time_parameter}
\end{figure}

\subsection{Analysis on training datasets}\label{subsec:traindata}
We investigate the influence of using different groups of the training datasets to the performance on the benchmarks.
As we mentioned in \S\ref{synthetic_data}, prior works used different sets of the training datasets and left uncertainties as to the contributions of their models to improvements. 
To unpack this issue, we examined the accuracy of our best model from \S\ref{subsec:tradeoff} with different settings of the training dataset. 
We obtained 80.0\% total accuracy by using only MJSynth, 75.6\% by using only SynthText, and 84.1\% by using both.
The combination of MJSynth and SynthText improved accuracy by more than 4.1\%, over the individual usages of MJSynth and SynthText. 
A lesson from this study is that the performance results using different training datasets are incomparable, and such comparisons fail to prove the contribution of the model, which is why we trained all models with the same training dataset, unless mentioned otherwise.

Interestingly, training on 20\% of MJSynth (1.8M) and 20\% of SynthText (1.1M) together (total 2.9M $\approx$ the half of SynthText) provides 81.3\% accuracy -- better performance than the individual usages of MJSynth or SynthText.
MJSynth and SynthText have different properties because they were generated with different options, such as distortion and blur.
This result showed that the diversity of training data can be more important than the number of training examples, and that the effects of using different training datasets is more complex than simply concluding more is better. 

\subsection{Analysis of trade-offs for module combinations}
\label{subsec:tradeoff}
Here, we focus on the accuracy-speed and accuracy-memory trade-offs shown in different combinations of modules.
We provide the full table of results in the supplementary materials.
See Figure~\ref{tradeoff} for the trade-off plots of all 24 combinations, including the six previously proposed STR models (Stars in Figure~\ref{tradeoff}).
In terms of the accuracy-time trade-off, Rosetta and STAR-net are on the frontier and the other four prior models are inside of the frontier.
In terms of the accuracy-memory trade-off, R2AM is on the frontier and the other five of previously proposed models are inside of the frontier.
Module combinations along the trade-off frontiers are labeled in ascending ascending order of accuracy (T1 to T5 for accuracy-time and P1 to P5 for accuracy-memory).

\paragraph{Analysis of combinations along the trade-off frontiers.}
As shown in Table~\ref{tab:trade-off-time}, T1 takes the minimum time by not including any transformation or sequential module. 
Moving from T1 to T5, the following modules are introduced in order (indicated as bold): ResNet, BiLSTM, TPS, and Attn. 
Note that from T1 to T5, a single module changes at a time. Our framework provides a smooth shift of methods that gives the least performance trade-off depending on the application scenario. 
They sequentially increase the complexity of the overall STR model, resulting in increased performance at the cost of computational efficiency.
ResNet, BiLSTM, and TPS introduce relatively moderate overall slow down (1.3ms$\rightarrow$10.9ms), while greatly boosting accuracy (69.5\%$\rightarrow$82.9\%).
The final change, Attn, on the other hand, only improves the accuracy by 1.1\% at a huge cost in efficiency (27.6 ms).

As for the accuracy-memory trade-offs shown in Table~\ref{tab:trade-off-memory}, P1 is the model with the least amount of memory consumption, and from P1 to P5 the trade-off between memory and accuracy takes place.
As in the accuracy-speed trade-off, we observe a single module shift at each step up to P5, where the changed modules are: Attn, TPS, BiLSTM, and ResNet.
They sequentially increase the accuracy at the cost of memory.
Compared to VGG used in T1, we observe that RCNN in P1-P4 is lighter and gives a good accuracy-memory trade-off. RCNN requires a small number of unique CNN layers that are repeatedly applied.
We observe that transformation, sequential, and prediction modules are not significantly contributing to the memory consumption (1.9M$\rightarrow$7.2M parameters).
While being lightweight overall, these modules provide accuracy improvements (75.4\%$\rightarrow$82.3\%).
The final change, ResNet, on the other hand, increases the accuracy by 1.7\% at the cost of increased memory consumption from 7.2M to 49.6M floating point parameters.
Thus, a practitioner concerned about memory consumption can be assured to choose specialized transformation, sequential, and prediction modules relatively freely, but should refrain from the use of heavy feature extractors like ResNets.

\paragraph{The most important modules for speed and memory.}
We have identified the module-wise impact on speed and memory by color-coding the scatter plots in Figure~\ref{tradeoff} according to module choices.
The full set of color-coded plots is in the supplementary materials.
Here, we show the scatter plots with the most speed- and memory-critical modules, namely the prediction and feature extraction modules, respectively, in Figure~\ref{acc_time_parameter}.

There are clear clusters of combinations according to the prediction and feature modules.
In the accuracy-speed trade-off, we identify CTC and Attn clusters (the addition of Attn significantly slows the overall STR model).
On the other hand, for accuracy-memory trade-off, we observe that the feature extractor contributes towards memory most significantly.
It is important to recognize that the most significant modules for each criteria differ, therefore, practitioners under different applications scenarios and constraints should look into different module combinations for the best trade-offs depending on their needs.

\begin{figure*}[t]

  \begin{subfigure}{0.16\textwidth} \centering
  	 \includegraphics[width=0.97\textwidth, height=0.10\textheight]{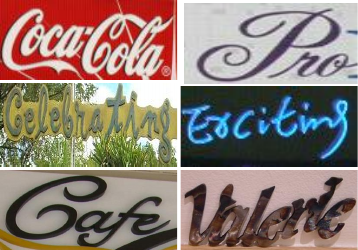}
     \caption{Difficult fonts.}\label{failure:figA}
   \end{subfigure}
   \begin{subfigure}{0.16\textwidth} \centering
     \includegraphics[width=0.97\textwidth, height=0.10\textheight]{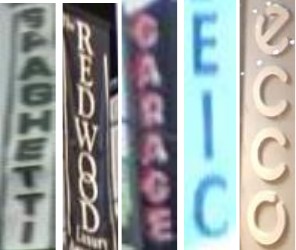}
     \caption{Vertical texts.}\label{failure:figB}
   \end{subfigure}
   \begin{subfigure}{0.16\textwidth} \centering
     \includegraphics[width=0.97\textwidth, height=0.10\textheight]{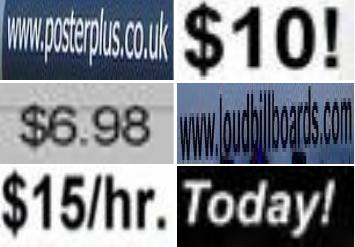}
     \caption{Special characters.}\label{failure:figC}
   \end{subfigure}
   \begin{subfigure}{0.16\textwidth} \centering
     \includegraphics[width=0.97\textwidth, height=0.10\textheight]{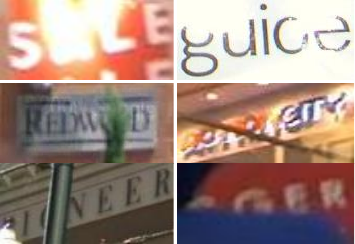}
     \caption{Occlusion.}\label{failure:figD}
   \end{subfigure}
   \begin{subfigure}{0.16\textwidth} \centering
     \includegraphics[width=0.97\textwidth, height=0.10\textheight]{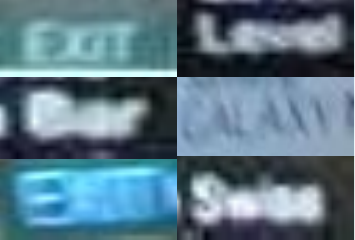}
     \caption{Low resolution.}\label{failure:figE}
   \end{subfigure}
   \begin{subfigure}{0.16\textwidth} \centering
     \includegraphics[width=0.97\textwidth, height=0.10\textheight]{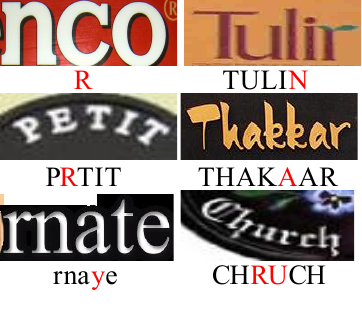}
     \caption{Label noise.}\label{failure:figE}
   \end{subfigure}
  \caption{Samples of failure cases on all combinations of our framework.}
  \label{failure}
\end{figure*}

\subsection{Module analysis}

\begin{table}[!htbp]
\small
	\begin{center}
      \begin{adjustbox}{width=0.475\textwidth}
		\begin{tabular}{cl|cc|r|r}
			\hline
            \multirow{2}{*}{Stage} & \multirow{2}{*}{Module} & \multicolumn{2}{c|}{Accuracy} & Time & params \cr 
            & & Regular (\%) & Irregular (\%) & ms/image & $\times10^6$ \cr \hline
            
            \multirow{2}{*}{\textbf{Trans.}} & None & 85.6 & 65.7 & N/A & N/A \cr 
                                            & TPS  & 86.7(\textbf{+1.1}) & 69.1(\textbf{+3.4})& \textbf{3.6} & \textbf{1.7} \cr \hline
            \multirow{3}{*}{\textbf{Feat.}} & VGG & 84.5 & 63.9 & 1.0 & 5.6 \cr
                                            & RCNN & 86.2(\textbf{+1.7}) & 67.3(\textbf{+3.4}) & \textbf{6.9} & 1.8 \cr
                                            & ResNet & 88.3(\textbf{+3.8}) & 71.0(\textbf{+7.1}) & 4.1 & \textbf{44.3} \cr \hline
            \multirow{2}{*}{\textbf{Seq.}} & None & 85.1 & 65.2 & N/A & N/A \cr 
                                    & BiLSTM  & 87.6(\textbf{+2.5}) & 69.7(\textbf{+4.5}) & \textbf{3.1} & \textbf{2.7} \cr \hline
            \multirow{2}{*}{\textbf{Pred.}} & CTC & 85.5 & 66.1 & 0.1 & 0.0 \cr 
                                    & Attn & 87.2(\textbf{+1.7}) & 68.7(\textbf{+2.6}) & \textbf{17.1} & \textbf{0.9} \cr \hline
		\end{tabular}
    \end{adjustbox}
    \end{center}
    \caption{Study of modules at the four stages with respect to total accuracy, inference time, and the number of parameters. The accuracies are acquired by taking the mean of the results of the combinations including that module. The inference time and the number of parameters are measured individually.}
	\label{result_module}
\end{table}

Here, we investigate the module-wise performances in terms of accuracy, speed, and memory demand. For this analysis, the marginalized accuracy of each module is calculated by averaging out the combination including the module in Table~\ref{result_module}.
Upgrading a module at each stage requires additional resources, time or memory, but provides performance improvements. The table shows that the performance improvement in irregular datasets is about two times that of regular benchmarks over all stages.
when comparing accuracy improvement versus time usage, a sequence of ResNet, BiLSTM, TPS, and Attn is the most efficient upgrade order of the modules from a base combination of None-VGG-None-CTC.
This order is the same order of combinations for the accuracy-time frontiers (T1$\rightarrow$T5). 
On the other hand, an accuracy-memory perspective finds RCNN, Attn, TPS, BiLSTM and ResNet as the most efficient upgrading order for the modules, like the order of the accuracy-memory frontiers (P1$\rightarrow$P5). 
Interestingly, the efficient order of modules for time is reverse from those for memory. The different properties of modules provide different choices in practical applications.
In addition, the module ranks in the two perspectives are the same as the order of the frontier module changes, and this shows that each module contributes to the performances similarly under all combinations.

\begin{figure}[t]
  \includegraphics[width=0.475\textwidth]{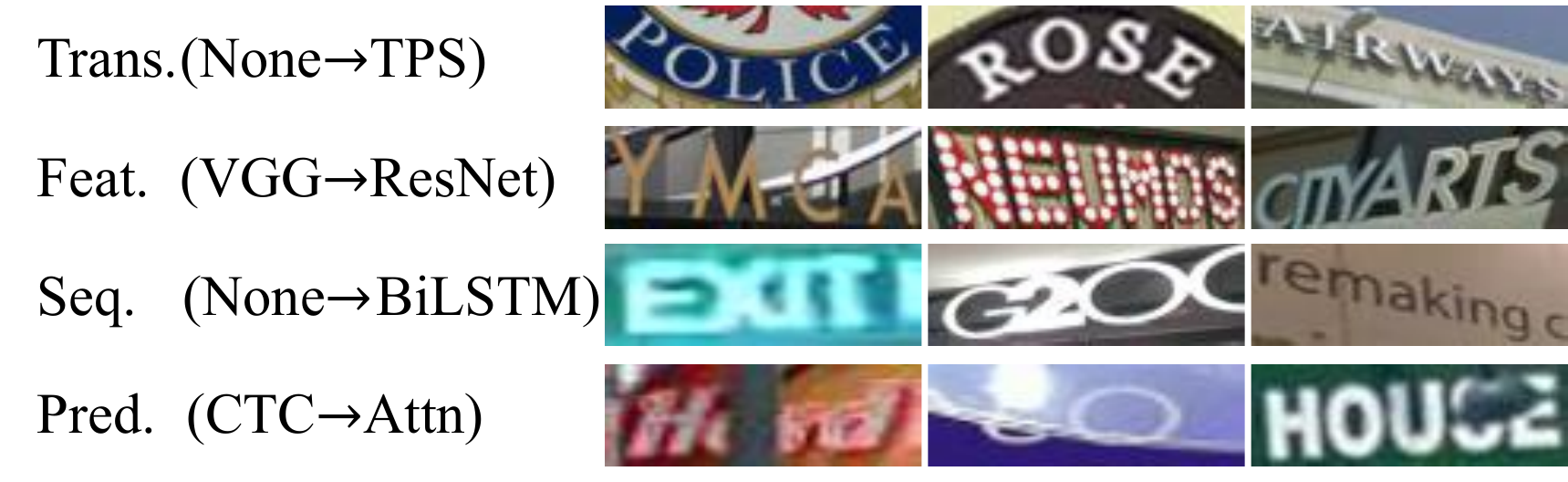}
  \caption{Challenging examples for the STR combinations without a specific module. All STR combinations without the notated modules failed to recognize text in the examples, but upgrading the module solved the problem.}
  \label{fig:benchmark_examples_module}
\end{figure}

\noindent\textbf{Qualitative analysis}
\label{subsec:qualitative}
Each module contributes to identify text by solving targeted difficulties of STR tasks, as described in \S\ref{model}.
Figure~\ref{fig:benchmark_examples_module} shows samples that are only correctly recognized when certain modules are upgraded (e.g. from VGG to ResNet backbone). Each row shows a module upgrade at each stage of our framework. Presented samples are failed before the upgrade, but becomes recognizable afterward. TPS transformation normalizes curved and perspective texts into a standardized view. Predicted results show dramatic improvements especially for ``POLICE'' in a circled brand logo and ``AIRWAYS'' in a perspective view of a storefront sign. 
Advanced feature extractor, ResNet, results in better representation power, improving on cases with heavy background clutter ``YMCA'', ``CITYARTS'') and unseen fonts (``NEUMOS'').
BiLSTM leads to better context modeling by adjusting the receptive field; it can ignore unrelatedly cropped characters (``I'' at the end of ``EXIT'', ``C'' at the end of ``G20''). 
Attention including implicit character-level language modeling finds missing or occluded character, such as ``a'' in ``Hard'', ``t'' in ``to'', and ``S'' in ``HOUSE''. 
These examples provide glimpses to the contribution points of the modules in real-world applications.  

\subsection{Failure case analysis} 
\label{subsec:failure}
We investigate failure cases of all 24 combinations.
As our framework derived from commonalities among proposed STR models, and our best model showed competitive performance with previously proposed STR models, the presented failure cases constitute a common challenge for the field as a whole.
We hope our analysis inspires future works in STR to consider addressing those challenges.

Among 8,539 examples in the benchmark datasets (\S\ref{dataset}), 644 images (7.5\%) are not correctly recognized by \emph{any} of the 24 models considered. We found six common failure cases as shown in Figure~\ref{failure}. The followings are discussion about the challenges of the cases and suggestion future research directions. 

\noindent \textbf{Calligraphic fonts:} font styles for brands, such as ``Coca Cola'', or shop names on streets, such as ``Cafe'', are still in remaining challenges. Such diverse expression of characters requires a novel feature extractor providing generalized visual features. Another possible approach is regularization because the model might be over-fitting to the font styles in a training dataset. \\
 \textbf{Vertical texts:} most of current STR models assumes horizontal text images, and thus structurally could not deal with vertical texts. Some STR models~\cite{yang2017learning,cheng2018aon} exploit vertical information also, however, vertical texts are not clearly covered yet. Further research would be needed to cover vertical texts.\\
 \textbf{Special characters:} since current benchmarks do not evaluate special characters, existing works exclude them during training. 
 This results in failure prediction, misleading 
 the model to treat them as alphanumeric characters. We suggest training \textit{with} special characters. This has resulted in a boost from 87.9\% to 90.3\% accuracy on IIIT. \\
 \textbf{Heavy occlusions:} current methods do not extensively exploit contextual information to overcome occlusion. Future researches may consider superior language models to maximally utilize context. \\
 \textbf{Low resolution:} existing models do not explicitly handle low resolution cases; image pyramids or super-resolution modules may improve performance. \\
 \textbf{Label noise:} We found some noisy (incorrect) labels in the failure examples. We examined all examples in the benchmark to identify the ratio of the noisy labels. All benchmark datasets contain noisy labels and the ratio of mislabeling without considering special character was 1.3\%, mislabeling with considering special character was 6.1\%, and mislabeling with considering case-sensitivity was 24.1\%. 

We make all failure cases available in our Github repository, hoping that they will inspire further researches on corner cases of the STR problem.
\section{Conclusion} \label{conclusion}

While there has been great advances on novel scene text recognition (STR) models, they have been compared on inconsistent benchmarks, leading to difficulties in determining whether and how a proposed module improves the STR baseline model.
This work analyzes the contribution of the existing STR models that was hindered under inconsistent experiment settings before. To achieve this goal, we have introduced a common framework among key STR methods, as well as consistent datasets: seven benchmark evaluation datasets and two training datasets (MJ and ST).
We have provided a fair comparison among the key STR methods compared, and have analyzed which modules brings about the greatest accuracy, speed, and size gains. We have also provided extensive analyses on module-wise contributions to typical challenges in STR as well as the remaining failure cases.

\section*{Acknowledgement}
The authors would like to thank Jaeheung Surh for helpful discussions.

{\small
\bibliographystyle{ieee_fullname}
\bibliography{main}
}

\clearpage
\appendix
\section{Contents}\label{sup:contents}

\noindent Appendix~\ref{sup:data-example} : \textbf{Dataset Matters in STR - examples} 
  \begin{itemize}
  \item We illustrate examples of the problematic datasets described in \S\ref{dataset} and \S\ref{implementation}.
  \end{itemize}
\noindent Appendix~\ref{sup:verification} : \textbf{STR Framework - verification} 
  \begin{itemize}
  \item We verify our STR module implementations for our framework by reproducing four existing STR models, namely CRNN~\cite{CRNN}, RARE~\cite{RARE}, GRCNN~\cite{GRCNN}, and FAN (w/o Focus Net)~\cite{FAN}.
  \end{itemize}
\noindent Appendix~\ref{sup:architecture} : \textbf{STR Framework - architectural details}
  \begin{itemize}
  \item 
  We describe the architectural details of all modules in our framework described in \S\ref{model}.
  \end{itemize}
\noindent Appendix~\ref{sup:experiments} : \textbf{STR Framework - full experimental results}
  \begin{itemize}
  \item We provide the comprehensive results of our experiments described in \S\ref{sec:experiments}, and discuss them in detail.
  \end{itemize}

\noindent Appendix~\ref{sup:additional_experiments} : \textbf{Additional Experiments}
  \begin{itemize}
  \item We provide 3 experiments; fine-tuning, varying training dataset size and test on COCO dataset~\cite{COCO}.
  \end{itemize}

\section{Dataset Matters in STR - examples}\label{sup:data-example}
\begin{figure}
\includegraphics[width=0.45\textwidth]{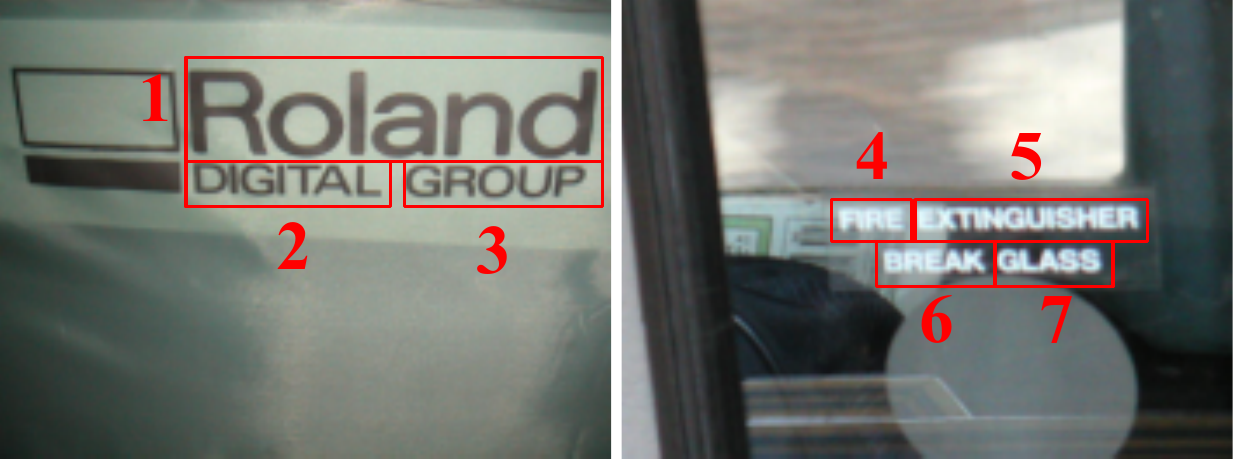}
	\centering
  \caption{7 missing examples of IC03-860 evaluation dataset. The examples represented by the red-colored word boxes of the two real scene images are included in IC03-867, but not in IC03-860. 
  }
  \label{missed_image}
\vspace{-2mm}
\end{figure}

\noindent\textbf{IC03 - 7 missing word boxes in 860 evaluation dataset.} 
The original IC03 evaluation dataset has 1,110 images, but prior works have conducted additional filtering, as described in \S\ref{dataset}. 
All papers have ignored all words that are either too short (less than 3 characters) or ones that contain non-alphanumeric characters.
Although all papers have supposedly applied the same data filtering method and should have reduced the evaluation set from 1,110 images to 867 images,
the reported example numbers are different: either the expected 867 images or a further reduced 860 images. We identified the missing examples as shown in Figure~\ref{missed_image}.

\begin{figure}
\includegraphics[width=0.45\textwidth]{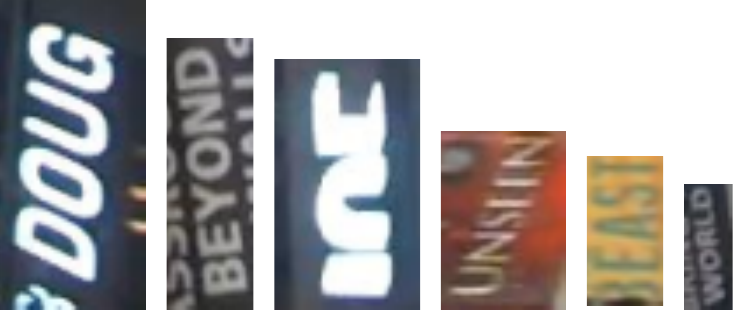}
	\centering
  \caption{Images that were filtered out of the IC15 evaluation dataset.}
  \label{IC15_rotate}
\vspace{-2mm}
\end{figure}

\noindent\textbf{IC15 - Filtered examples in evaluation dataset.}
The IC15 dataset originally contains 2,077 examples for its evaluation set, however prior works~\cite{FAN, Bai_2018_CVPR} have filtered it down to 1,811 examples and have not given unambiguous specifications between them for deciding on which example to discard.
To resolve this ambiguity, we have contacted one of the authors, who shared the specific dataset used for the evaluation.
This information is made available with the source code on Github.
A few sample images that have been filtered out of the IC15 evaluation dataset is shown in Figure~\ref{IC15_rotate}.

\begin{figure}
\includegraphics[width=0.45\textwidth]{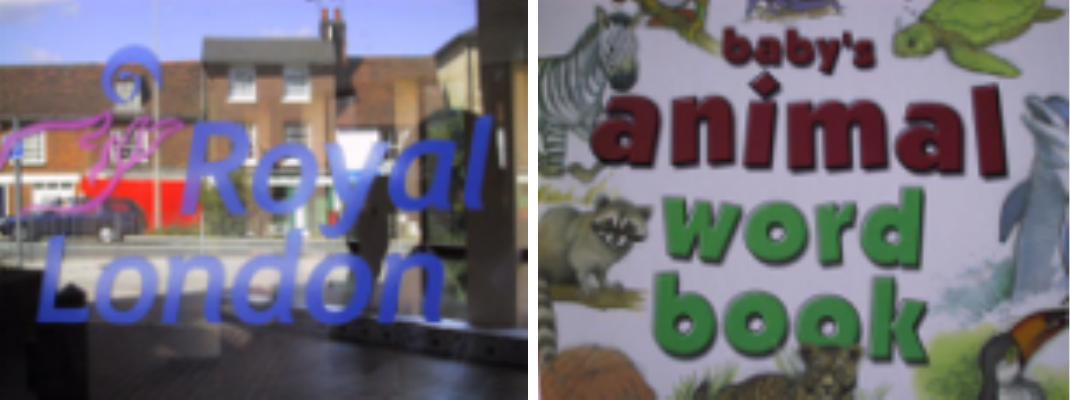}
	\centering
  \caption{
    Duplicated scene images. These are example images that have been found in both the IC03 training dataset and the IC13 evaluation dataset.
  }
  \label{intersection}
\vspace{-2mm}
\end{figure}

\noindent\textbf{IC03 and IC13 - Duplicated images between IC03 training dataset and IC13 evaluation dataset.}
Figure~\ref{intersection} shows two images from the subset given by the intersection between the IC03 training dataset and the IC13 evaluation dataset. In our investigation, a total of 34 duplicated scene images have been found, amounting to 215 duplicate word boxes, in total. Therefore, when one assesses the performance of a model on the IC13 evaluation data, he/she should be mindful of these overlapping data.

\begin{table*}[t]
	\begin{center}
        \begin{adjustbox}{width=0.975\textwidth}
		\begin{tabular}{lllccccccccccccc}
			\hline
			\multirow{2}{*}{Model} & \multirow{2}{*}{} & \multirow{2}{*}{Train Data} &
			IIIT & SVT & \multicolumn{2}{c}{IC03} & \multicolumn{2}{c}{IC13} & \multicolumn{2}{c}{IC15} & SP & CT \cr
			&&& 3000 & 647 & 860 & 867 & 857 & 1015 & 1811 & 2077 & 645 & 288  \cr\hline
            CRNN \cite{CRNN}     & reported in paper  & MJ   &78.2 & 80.8 & 89.4& $-$& $-$ & 86.7 & $-$ & $-$ & $-$ & $-$ \cr
            CRNN & our implementation & MJ  & 81.2 & 81.8 & 91.1 & $-$& $-$ & 87.6 & $-$ & $-$ & $-$ & $-$\cr
            RARE  \cite{RARE}     & reported in paper & MJ       &81.9 & 81.9 & 90.1 & $-$ & 88.6 & $-$ & $-$ & $-$ & 71.8 & 59.2  \cr
            RARE & our implementation & MJ & 83.1 & 84.0 & 92.2 & $-$ & 91.3 & $-$ & $-$ & $-$  & 74.3 & 64.2\cr
			   GRCNN  \cite{GRCNN}   & reported in paper & MJ      & 80.8 & 81.5 & 91.2 & $-$  & $-$ & $-$ & $-$&  $-$& $-$ & $-$  \cr     
            GRCNN & our implementation & MJ &  82.0 & 81.1 & 90.3 & $-$ & $-$ & $-$& $-$&  $-$& $-$ & $-$\cr
            FAN (w/o Focus Net)\cite{FAN} & reported in paper & MJ+ST & 83.7 & 82.2 & $-$ & 91.5 & $-$ & 89.4 & 63.3 & $-$ & $-$ & $-$ \cr	           
            FAN (w/o Focus Net) & our implementation & MJ+ST & 86.4 & 86.2  & $-$ & 94.3 & $-$ & 90.6 & 73.3 & $-$ & $-$ & $-$\cr		
            \hline
        \end{tabular}
        \end{adjustbox}       
	\end{center}
        \captionof{table}{Sanity checking our experimental platform by reproducing the existing four STR models: CRNN~\cite{CRNN}, RARE~\cite{RARE}, GRCNN~\cite{GRCNN} and FAN (w/o Focus Net~\cite{FAN}).}
    \label{sup:sanity}
\end{table*}

\section{STR Framework - verification}\label{sup:verification}
To show the correctness of our implemented module for our framework, we reproduce the performances of existing models that can be re-built by our framework. Specifically, we compare the results of our implementation of CRNN, RARE, GRCNN, and FAN (w/o Focus Net)~\cite{CRNN,RARE,GRCNN,FAN} from those of publicly reported by the authors. We implemented each module as described in their original papers, and also we followed the training and evaluation pipelines of their original papers to train the individual models.
Table~\ref{sup:sanity} shows the results. Our implementation has overall similar performance with reported result in their paper, which verify the sanity of our implementations and experiments.

\section{STR Framework - architectural details}\label{sup:architecture}
In this appendix, we describe each module of our framework in terms of its concept and architectural specifications. We first introduce common notations used in this appendix and then explain the modules of each stage; Trans., Feat., Seq., and Pred.

\noindent \textbf{Notations} For a simple expression for a neural network architecture, we denote `c', `k', `s' and `p' for the number of the output channel, the size of kernel, the stride, and the padding size respectively. BN, Pool, and FC denote the batch normalization layer, the max pooling layer, and the fully connected layer, respectively. In the case of convolution operation with the stride of 1 and the padding size of 1, `s' and `p' are omitted for convenience.

\subsection{Transformation stage}
The module of this stage transforms the input image $X$ into the normalized image $\tilde{X}$.
We explained the concept of TPS~\cite{RARE,liu2016star} in \S\ref{subsec:transformation}, but here we deliver its mathematical background and the implementation details.   

\noindent\textbf{TPS transformation}: TPS generates a normalized image that shows a focused region of an input image. To build this pipeline, TPS consists of a sequence of processes; finding a text boundary, linking the location of the pixels in the boundary to those of the normalized image, and generating a normalized image by using the values of pixels and the linking information. Such processes are called as localization network, grid generator, and image sampler, respectively. Conceptually, TPS employs a smooth spline interpolation between a set of $F$ fiducial points that represented a focused boundary of text in an image. Here, $F$ indicates the constant number of fiducial points. 

The localization network explicitly calculates $x,y$-coordinates of $F$ fiducial points on an input image,  $X$.
The coordinates are denoted by $\mathbf{C} = \left[ \mathbf{c}_{1}, \dots, \mathbf{c}_{F} \right] \in \mathbb{R}^{2 \times F}$, whose $f$-th column $\mathbf{c}_{f}=\left[x_{f},y_{f}\right]^{\intercal}$ contains the coordinates of the $f$-th fiducial point. $\tilde{\mathbf{C}}$ represents pre-defined top and bottom locations on the normalized image, $\tilde{X}$.

The grid generator provides a mapping function from the identified regions by the localization network to the normalized image. The mapping function can be parameterized by a matrix $\mathbf{T}\in\mathbb{R}^{2\times(F+3)}$, which is computed by
\begin{equation}
  \mathbf{T}=\left(\boldsymbol{\Delta}_{\tilde{\mathbf{C}}}^{-1}\left[\begin{array}{c}
    \mathbf{C}^{\intercal}\\
    \mathbf{0}^{3\times2}
    \end{array}
  \right]\right)^{\intercal}
  \label{eq:tps-solve}
\end{equation}
where $\boldsymbol{\Delta}_{\mathbf{C}'}\in\mathbb{R}^{(F+3)\times(F+3)}$ is a matrix determined only by $\tilde{\mathbf{C}}$, thus also a constant:
\begin{equation}
\boldsymbol{\Delta}_{\tilde{\mathbf{C}}}=\begin{bmatrix}\mathbf{1}^{F\times1} & \tilde{\mathbf{C}}^{\intercal} & \mathbf{R}\\
\mathbf{0} & \mathbf{0} & \mathbf{1}^{1\times F}\\
\mathbf{0} & \mathbf{0} & \tilde{\mathbf{C}}
\end{bmatrix}
\end{equation}
where the element of $i$-th row and $j$-th column of $\mathbf{R}$ is $d_{ij}^{2} \ln d_{ij}$, $d_{ij}$ is the euclidean distance between $\tilde{c}_i$ and $\tilde{c}_j$. The pixels of grid on the normalized image $\tilde{X}$ is denoted by $\tilde{P}=\left\{ \tilde{p}_i \right\}_{i=1,\dots,N}$
, where $\tilde{p}_i=\left[\tilde{x}_i,\tilde{y}_i\right]^{\intercal}$ is the x,y-coordinates of the $i$-th pixel, $N$ is the number of pixels. For every pixel $\tilde{p}_i$ on $\tilde{X}$, we find the corresponding point $p_i=\left[x_i,y_i\right]^{\intercal}$ on $X$, by applying the transformation:
\begin{align}
  p_i & = \mathbf{T}\left[1,\tilde{x}_i,\tilde{y}_i,r_{i1},\dots,r_{iF}\right]^{\intercal}\label{eq:tps-trans}\\
  r_{if} & = d_{if}^{2}\ln d_{if}
\end{align}
where $d_{if}$ is the euclidean distance between pixel $\tilde{p}_i$ and the $f$-th base fiducial point $\tilde{c}_f$.
By iterating Eq.~\ref{eq:tps-trans} over all points in ${\cal \tilde{P}}$, we generate a grid ${\cal P}=\left\{ p_i\right\}_{i=1,\dots,N}$ on the input image $X$. 

Finally, the image sampler produces the normalized image by interpolating the pixels in the input images which are determined by the grid generator.

\noindent\textbf{TPS-Implementation:} TPS requires the localization network calculating fiducial points of an input image. 
We designed the localization network by following most of the components of prior work~\cite{RARE}, and added batch normalization layers and adaptive average pooling to stabilize the training of the network.
Table~\ref{tab:localizationnetwork} shows the details of our architecture.
In our implementation, the localization network has 4 convolution layers, 
each followed by a batch normalization layer and 2 x 2 max-pooling layer. 
The filter size, padding size, and stride are 3, 1, 1 respectively, for all convolutional layers. 
Following the last convolutional layer is an adaptive average pooling layer (APool in Table~\ref{tab:localizationnetwork}).
After that, two fully connected layers are following: 512 to 256 and 256 to 2F. 
Final output is 2F dimensional vector which corresponds to the value of $x,y$-coordinates of $F$ fiducial points on input image.
Activation functions for all layers are the ReLU.

\begin{table}
\begin{centering}
\begin{adjustbox}{width=0.40\textwidth}
\begin{tabular}[t]{|l|lc|c|}
\hline 
\textbf{Layers} & \multicolumn{2}{c|}{\textbf{Configurations}} & \textbf{Output}\\
\hline 
Input & \multicolumn{2}{c|}{grayscale image} & $100\times 32$\\
\hline 
Conv1 & c: $64$ & k: $3\times3$ & $100\times 32$\\
\hline
BN1 & \multicolumn{2}{c|}{-} &$100\times 32$\\
\hline
Pool1 & k: $2\times2$ & s: $2\times2$  & $50\times 16$\\
\hline 
Conv2 & c: $128$ & k: $3\times3$ & $50\times 16$\\
\hline
BN2 & \multicolumn{2}{c|}{-} &$50\times 16$\\
\hline
Pool2 & k: $2\times2$ & s: $2\times2$ & $25\times 8$\\
\hline 
Conv3 & c: $256$ & k: $3\times3$ & $25\times 8$\\
\hline
BN3 & \multicolumn{2}{c|}{-} &$25\times 8$\\
\hline
Pool3 & k: $2\times2$ & s: $2\times2$ & $12\times 4$\\ 
\hline 
Conv4 & c: $512$ & k: $3\times3$ & $12\times 4$\\
\hline
BN4 & \multicolumn{2}{c|}{-} &$12\times 4$\\
\hline
APool & \multicolumn{2}{c|}{$512\times12\times 4 \rightarrow 512\times1$} & $512$\\
\hline 
FC1 & \multicolumn{2}{c|}{$512 \rightarrow 256$} & $256$\\
\hline 
FC2 & \multicolumn{2}{c|}{$256 \rightarrow 2F$} & $2F$\\
\hline 
\end{tabular}
\end{adjustbox}
\par\end{centering}
\caption{Architecture of the localization network in TPS. The localization network extracts the location of the text line, that is, the $x$- and $y$-coordinates of the fiducial points $F$ within the input image.}
\label{tab:localizationnetwork}
\end{table}

\begin{table}
\begin{centering}
\begin{adjustbox}{width=0.40\textwidth}
\begin{tabular}[t]{|l|lc|c|}
\hline 
\textbf{Layers} & \multicolumn{2}{c|}{\textbf{Configurations}} & \textbf{Output}\\
\hline 
Input & \multicolumn{2}{c|}{grayscale image} & $100\times 32$\\
\hline 
Conv1 & c: $64$ & k: $3\times3$ & $100\times 32$\\
\hline
Pool1 & k: $2\times2$ & s: $2\times2$ & $50\times 16$\\
\hline 
Conv2 & c: $128$ & k: $3\times3$ & $50\times 16$\\
\hline
Pool2 & k: $2\times2$ & s: $2\times2$ & $25\times 8$\\
\hline 
Conv3 & c: $256$ & k: $3\times3$ & $25\times 8$ \\
\hline
Conv4 & c: $256$ & k: $3\times3$ & $25\times 8$ \\ 
\hline
Pool3 & k: $1\times2$ & s: $1\times2$ & $25\times 4$\\
\hline 
Conv5 & c: $512$ & k: $3\times3$ & $25\times 4$ \\
\hline
BN1 & \multicolumn{2}{c|}{-} &$25\times 4$\\
\hline
Conv6 & c: $512$ & k: $3\times3$ & $25\times 4$\\
\hline
BN2 & \multicolumn{2}{c|}{-} &$25\times 4$\\
\hline
Pool4 & k: $1\times2$ & s: $1\times2$ & $25\times 2$\\
\hline 
\multirow{2}{*}{Conv7} & c: $512$ & k: $2\times2$ & \multirow{2}{*}{$24\times 1$}\\
 & s: $1\times1$ & p: $0\times0$ & \\
\hline
\end{tabular}
\end{adjustbox}
\par\end{centering}
\caption{Architecture of VGG.}
\label{tab:vgg}
\end{table}

\begin{table}
\begin{centering}
\begin{adjustbox}{width=0.40\textwidth}
\begin{tabular}[t]{|l|lc|c|}
\hline 
\textbf{Layers} & \multicolumn{2}{c|}{\textbf{Configurations}} & \textbf{Output}\\
\hline 
Input & \multicolumn{2}{c|}{grayscale image} & $100\times 32$\\
\hline 
Conv1 & c: $64$ & k: $3\times3$ & $100\times 32$\\
\hline
Pool1 & k: $2\times2$ & s: $2\times2$ & $50\times 16$\\
\hline 
GRCL1& \multicolumn{2}{c|}{$\begin{bmatrix}\rm{c:}64,\rm{k:}3\times3\end{bmatrix}\times 5$} & $50\times 16$\\
\hline 
Pool2 & k: $2\times2$ & s: $2\times2$ & $25\times 8$\\
\hline 
GRCL2& \multicolumn{2}{c|}{$\begin{bmatrix}\rm{c:}128,\rm{k:}3\times3\end{bmatrix}\times 5$} & $25\times 8$\\
\hline 
\multirow{2}{*}{Pool3} & k: $2\times2$ &  & \multirow{2}{*}{$26\times 4$}\\
 & s: $1\times2$ & p: $1\times0$ & \\
\hline 
GRCL3& \multicolumn{2}{c|}{$\begin{bmatrix}\rm{c:}256,\rm{k:}3\times3\end{bmatrix}\times 5$} & $26\times 4$\\
\hline 
\multirow{2}{*}{Pool4} & k: $2\times2$ &  & \multirow{2}{*}{$27\times 2$}\\
 & s: $1\times2$ & p: $1\times0$ & \\
\hline 
\multirow{2}{*}{Conv2} & c: $512$ & k: $3\times3$ & \multirow{2}{*}{$26\times 1$}\\
 & s: $1\times1$ & p: $0\times0$ & \\
\hline
\end{tabular}
\end{adjustbox}
\par\end{centering}
\caption{Architecture of RCNN.}
\label{tab:rcnn}
\end{table}

\begin{table}
\begin{centering}
\begin{adjustbox}{width=0.40\textwidth}
\begin{tabular}[t]{|l|lc|c|}
\hline 
\textbf{Layers} & \multicolumn{2}{c|}{\textbf{Configurations}} & \textbf{Output}\\
\hline 
Input & \multicolumn{2}{c|}{grayscale image} & $100\times 32$\\
\hline 
Conv1 & c: $32$ & k: $3\times3$ & $100\times 32$\\
\hline
Conv2 & c: $64$ & k: $3\times3$ & $100\times 32$\\
\hline
Pool1 & k: $2\times2$ & s: $2\times2$ & $50\times 16$\\
\hline 
Block1& \multicolumn{2}{c|}{$\begin{bmatrix}\rm{c:}128,\rm{k:}3\times3\\\rm{c:}128,\rm{k:}3\times3\end{bmatrix}\times 1$} & $50\times 16$\\
\hline 
Conv3 & c: $128$ & k: $3\times3$ & $50\times 16$\\
\hline
Pool2 & k: $2\times2$ & s: $2\times2$ & $25\times 8$\\
\hline 
Block2& \multicolumn{2}{c|}{$\begin{bmatrix}\rm{c:}256,\rm{k:}3\times3\\\rm{c:}256,\rm{k:}3\times3\end{bmatrix}\times 2$} & $25\times 8$\\
\hline 
Conv4 & c: $256$ & k: $3\times3$ & $25\times 8$\\
\hline
\multirow{2}{*}{Pool3} & k: $2\times2$ &  & \multirow{2}{*}{$26\times 4$}\\
 & s: $1\times2$ & p: $1\times0$ & \\
\hline 
Block3& \multicolumn{2}{c|}{$\begin{bmatrix}\rm{c:}512,\rm{k:}3\times3\\\rm{c:}256,\rm{k:}3\times3\end{bmatrix}\times 5$} & $26\times 4$\\
\hline 
Conv5 & c: $512$ & k: $3\times3$ & $26\times 4$\\
\hline
Block4& \multicolumn{2}{c|}{$\begin{bmatrix}\rm{c:}512,\rm{k:}3\times3\\\rm{c:}512,\rm{k:}3\times3\end{bmatrix}\times 3$} & $26\times 4$\\
\hline 
\multirow{2}{*}{Conv6} & c: $512$ & k: $2\times2$ & \multirow{2}{*}{$27\times 2$}\\
 & s: $1\times2$ & p: $1\times0$ & \\
\hline
\multirow{2}{*}{Conv7} & c: $512$ & k: $2\times2$ & \multirow{2}{*}{$26\times 1$}\\
 & s: $1\times1$ & p: $0\times0$ & \\
\hline 
\end{tabular}
\end{adjustbox}
\par\end{centering}
\caption{Architecture of ResNet.} 
\label{tab:ResNet}
\end{table}

\subsection{Feature extraction stage}
In this stage, a CNN abstract an input image (i.e., $X$ or $\tilde{X}$) and outputs a feature map $V = \{v_i\}, i=1,\dots,I$ ($I$ is the number of columns in the feature map).

\noindent\textbf{VGG}:
we implemented VGG~\cite{VGG} which is used in CRNN~\cite{CRNN} and RARE~\cite{RARE}.
We summarized the architecture in Table~\ref{tab:vgg}.
The output of VGG is 512 channels $\times$ 24 columns.

\noindent\textbf{Recurrently applied CNN (RCNN)}:
As a RCNN module, we implemented a Gated RCNN (GRCNN)~\cite{GRCNN} which is a variant of RCNN that can be applied recursively with a gating mechanism.
The architectural details of the module are shown in Table~\ref{tab:rcnn}.
The output of RCNN is 512 channels $\times$ 26 columns.

\noindent\textbf{Residual Network (ResNet)}:
As a ResNet~\cite{ResNet} module, we implemented the same network which is used in FAN~\cite{FAN}.
It has 29 trainable layers in total.
The details of the network is shown in Table~\ref{tab:ResNet}.
The output of ResNet is 512 channels $\times$ 26 columns.

\subsection{Sequence modeling stage}
Some previous works used Bidirectional LSTM \textbf{(BiLSTM)} to make a contextual sequence $H = \rm{Seq.}(V)$ after the Feat. stage~\cite{CRNN}.

\noindent \textbf{BiLSTM}:
We implemented 2-layers BiLSTM~\cite{BiLSTM} which is used in CRNN~\cite{CRNN}.
In the followings, we explain a BiLSTM layer used in our framework:
A $l^{\rm{th}}$ BiLSTM layer identifies two hidden states, $h_{i}^{(t),\rm{f}}$ and $h_{i}^{(t),\rm{b}}$ $\forall t$, calculated through time sequence and its reverse. 
Following \cite{CRNN}, we additionally applied a FC layer between BiLSTM layers to determine one hidden state, $\hat{h}_{t}^{(l)}$, by using the two identified hidden states, $h_{t}^{(l),\rm{f}}$ and $h_{t}^{(l),\rm{b}}$. 
The dimensions of all hidden states including the FC layer was set as $256$.

\noindent \textbf{None} indicates not to use any Seq. modules upon the output of the Feat. modules, that is, $H = V$.

\subsection{Prediction stage}
A prediction module produces the final prediction output from the input $H$, (i.e., $Y=y_1,y_2,\dots$), which is a sequence of characters.
We implemented two modules: Connectionist Temporal Classification (CTC)~\cite{CTC} based and Attention mechanism (Attn) based Pred. module.
In our experiments, we make the character label set $C$ which include 36 alphanumeric characters.
For the CTC, additional \textit{blank} token is added to the label set due to the characteristics of the CTC.
For the Attn, additional end of sentence (EOS) token is added to the label set due to the characteristics of the Attn.
That is, the number of character set $C$ is 37.

\noindent\textbf{Connectionist Temporal Classification (CTC)}: 
CTC takes a sequence $\mathbf{H}=h_{1},\dots,h_{T}$, where $T$ is the sequence length, and outputs the probability of $\pi$, which is defined as 
\begin{equation}
p(\pi|H)=\prod_{t=1}^{T}y_{\pi_{t}}^{t}
\end{equation}
where $y_{\pi_{t}}^{t}$ is the probability of generating character $\pi_{t}$ at each time step $t$.
After that, the mapping function $M$ which maps $\pi$ to $Y$ by removing repeated characters and blanks.
For instance, $M$ maps ``\texttt{aaa-{}-b-b-c-ccc-c-{}-}'' onto ``\texttt{abbccc}'', where '\texttt{-}' is \textit{blank} token.
The conditional probability is defined as the sum of probabilities of all $\pi$ that are mapped by $M$ onto $Y$, which is
\begin{equation}
p(Y|H)=\sum_{\pi:M(\pi)=Y}p(\pi|H)\label{eq:stringprob}
\end{equation}
At testing phase, the predicted label sequence is calculated by taking the highest probability character $\pi_{t}$ at each time
step $t$, and map the $\pi$ onto $Y$:
\begin{equation}
Y^{*}\approx M(\arg\max_{\pi}p(\pi|H))
\end{equation}

\noindent\textbf{Attention mechanism (Attn)}:
We implemented one layer LSTM attention decoder~\cite{NMT} which is used in FAN, AON, and EP~\cite{FAN,cheng2018aon,Bai_2018_CVPR}.
The details are as follows:
at $t$-step, the decoder predicts an output $y_t$ as
\begin{align}
y_t = \rm{softmax}(W_o s_t+b_o)
\end{align}
where $W_0$ and $b_0$ are trainable parameters. $s_t$ is the decoder LSTM hidden state at time $t$ as
\begin{align}
s_t = \rm{LSTM}(y_{t-1}, c_t, s_{t-1})
\end{align}
and $c_t$ is a context vector, which is computed as the weighted sum of $H=h_1, ...h_I$ from the former stage as
\begin{align}
c_t = \sum_{i=1}^{I} \alpha_{ti} h_i
\end{align}
where $\alpha_{ti}$ is called attention weight and computed by 
\begin{align}
	\alpha_{ti} = \frac{exp(e_{ti})}{\sum_{k=1}^I exp(e_{tk})}
\end{align}
where
\begin{align}
e_{ti} = v^\intercal \tanh(Ws_{t-1} + Vh_i + b)
\end{align}
and $v$, $W$, $V$ and $b$ are trainable parameters.
The dimension of LSTM hidden state was set as $256$.  

\subsection{Objective function}
Denote the training dataset by $TD=\{X_i,Y_i\}$,
where $X_i$ is the training image and $Y_i$ is the word label. 
The training conducted by minimizing the objective function that negative log-likelihood of the conditional probability of word label.
\begin{equation}
O=-\sum_{X_i,Y_i\in{TD}}\log p(Y_i|X_i)\label{eq:objective}
\end{equation}
This function calculates a cost from an image and its word label, and the modules in the framework are trained end-to-end manner.

\begin{table*}
	\centering\noindent
   	 \begin{adjustbox}{width=0.975\textwidth}
     \begin{threeparttable}
     \begin{tabular}{l|c|c|l|c|cccccccccc|crrr}
			\hline
  			\multirow{2}{*}{\#} & \multirow{2}{*}{Trans.} & \multirow{2}{*}{Feat.} & \multirow{2}{*}{Seq.} & \multirow{2}{*}{Pred.} &
			IIIT & SVT & \multicolumn{2}{c}{IC03} & \multicolumn{2}{c}{IC13} & \multicolumn{2}{c}{IC15} & SP & CT & Acc. & Time & params & FLOPS \cr
			&&&&& 3000 & 647 & 860& 867 & 857 & 1015 & 1811 & 2077 & 645 & 288 & Total & ms & $\times10^6$ & $\times10^9$ \cr\hline
1 & \multirow{12}{*}{None} & \multirow{4}{*}{VGG} & \multirow{2}{*}{None} & CTC & 76.2 & 73.8 & 86.7 & 86.0 & 84.8 & 81.9 & 56.6 & 52.4 & 56.6 & 49.9 & 69.5 & 1.3 & 5.6 & 1.2 \\
2 &  &  &  & Attn & 80.1 & 78.4 & 91.0 & 90.5 & 88.5 & 86.3 & 63.0 & 58.3 & 66.0 & 56.1 & 74.6 & 19.0 & 6.6 & 1.6 \\\cline{4-19}
3\tnote{1} &  &  & \multirow{2}{*}{BiLSTM} & CTC & 82.9 & 81.6 & 93.1 & 92.6 & 91.1 & 89.2 & 69.4 & 64.2 & 70.0 & 65.5 & 78.4 & 4.4 & 8.3 & 1.4\\
4 &  &  &  & Attn & 84.3 & 83.8 & 93.7 & 93.1 & 91.9 & 90.0 & 70.8 & 65.4 & 71.9 & 66.8 & 79.7 & 21.2 & 9.1 & 1.6\\\cline{3-19}
5 &  & \multirow{4}{*}{RCNN} &  \multirow{2}{*}{None} & CTC & 80.9 & 78.5 & 90.5 & 89.8 & 88.4 & 85.9 & 65.1 & 60.5 & 65.8 & 60.3 & 75.4 & 7.7 & 1.9 & 1.6\\
6\tnote{2} &  &  &  & Attn & 83.4 & 82.4 & 92.2 & 92.0 & 90.2 & 88.1 & 68.9 & 63.6 & 72.1 & 64.9 & 78.5 & 24.1 & 2.9 & 2.0\\\cline{4-19}
7\tnote{3} &  &  & \multirow{2}{*}{BiLSTM} & CTC & 84.2 & 83.7 & 93.5 & 93.0 & 90.9 & 88.8 & 71.4 & 65.8 & 73.6 & 68.1 & 79.8 & 10.7 & 4.6 & 1.8\\
8 &  &  &  & Attn & 85.7 & 84.8 & 93.9 & 93.4 & 91.6 & 89.6 & 72.7 & 67.1 & 75.0 & 69.2 & 81.0 & 27.4 & 5.5 & 2.0\\\cline{3-19}
9\tnote{4} &  & \multirow{4}{*}{ResNet} &  \multirow{2}{*}{None} & CTC & 84.3 & 84.7 & 93.4 & 92.9 & 90.9 & 89.0 & 71.2 & 66.0 & 73.8 & 69.2 & 80.0 & 4.7 & 44.3 & 10.1\\
10 &  &  &  & Attn & 86.1 & 85.7 & 94.0 & 93.6 & 91.9 & 90.1 & 73.5 & 68.0 & 74.5 & 72.2 & 81.5 & 22.2 & 45.3 & 10.5\\\cline{4-19}
11 &  &  & \multirow{2}{*}{BiLSTM} & CTC & 86.2 & 86.0 & 94.4 & 94.1 & 92.6 & 90.8 & 73.6 & 68.0 & 76.0 & 72.2 & 81.9 & 7.8 & 47.0 & 10.3\\
12 &  &  &  & Attn & 86.6 & 86.2 & 94.1 & 93.7 & 92.8 & 91.0 & 75.6 & 69.9 & 76.4 & 72.6 & 82.5 & 25.0 & 47.9 & 10.5\\
\hline
13 & \multirow{12}{*}{TPS} & \multirow{4}{*}{VGG} &  \multirow{2}{*}{None} & CTC & 80.0 & 78.0 & 90.1 & 89.7 & 88.7 & 87.5 & 65.1 & 60.6 & 65.5 & 57.0 & 75.1 & 4.8 & 7.3 & 1.6\\
14 &  &  &  & Attn & 82.9 & 82.3 & 92.0 & 91.7 & 90.5 & 89.2 & 69.4 & 64.2 & 73.0 & 62.2 & 78.5 & 21.0 & 8.3 & 2.0\\\cline{4-19}
15 &  &  & \multirow{2}{*}{BiLSTM} & CTC & 84.6 & 83.8 & 93.3 & 92.9 & 91.2 & 89.4 & 72.4 & 66.8 & 74.0 & 66.8 & 80.2 & 7.6 & 10.0 & 1.8\\
16\tnote{5} &  &  &  & Attn & 86.2 & 85.8 & 93.9 & 93.7 & 92.6 & 91.1 & 74.5 & 68.9 & 76.2 & 70.4 & 82.0 & 23.6 & 10.8 & 2.0\\\cline{3-19}
17 &  & \multirow{4}{*}{RCNN} &  \multirow{2}{*}{None} & CTC & 82.8 & 81.7 & 92.0 & 91.6 & 89.5 & 88.4 & 69.8 & 64.6 & 71.3 & 61.2 & 78.3 & 10.9 & 3.6 & 1.9\\
18 &  &  &  & Attn & 85.1 & 84.0 & 93.1 & 93.1 & 91.5 & 90.2 & 72.4 & 66.8 & 75.6 & 64.9 & 80.6 & 26.4 & 4.6 & 2.3\\\cline{4-19}
19 &  &  & \multirow{2}{*}{BiLSTM} & CTC & 85.1 & 84.3 & 93.5 & 93.1 & 91.4 & 89.6 & 73.4 & 67.7 & 74.4 & 69.1 & 80.8 & 14.1 & 6.3 & 2.1\\
20 &  &  &  & Attn & 86.3 & 85.7 & 94.0 & 94.0 & 92.8 & 91.1 & 75.0 & 69.2 & 77.7 & 70.1 & 82.3 & 30.1 & 7.1 & 2.3\\\cline{3-19}
21 &  & \multirow{4}{*}{ResNet} &  \multirow{2}{*}{None} & CTC & 85.0 & 85.7 & 94.0 & 93.6 & 92.5 & 90.8 & 74.6 & 68.8 & 75.2 & 71.0 & 81.5 & 8.3 & 46.0 & 10.5\\
22 &  &  &  & Attn & 87.1 & 87.1 & 94.3 & 93.9 & 93.2 & 91.8 & 76.5 & 70.6 & 78.9 & 73.2 & 83.3 & 25.6 & 47.0 & 10.9\\\cline{4-19}
23\tnote{6} &  &  & \multirow{2}{*}{BiLSTM} & CTC & 87.0 & 86.9 & 94.4 & 94.0 & 92.8 & 91.5 & 76.1 & 70.3 & 77.5 & 71.7 & 82.9 & 10.9 & 48.7 & 10.7\\
24\tnote{7} &   &   &   & Attn & \textbf{87.9} & \textbf{87.5} & \textbf{94.9} & \textbf{94.4} & \textbf{93.6} & \textbf{92.3} & \textbf{77.6} & \textbf{71.8} & \textbf{79.2} & \textbf{74.0} & \textbf{84.0} & 27.6 & 49.6 & 10.9\\
\hline
		\end{tabular}
		\begin{tablenotes}
            \item[]$^1$ CRNN. $^2$ R2AM. $^3$ GRCNN. $^4$ Rosetta. $^5$ RARE. $^6$ STAR-Net. $^7$ our best model.
  		\end{tablenotes}
        \end{threeparttable}
        \end{adjustbox}
     \captionof{table}{The full experimental results for all 24 STR combinations.
Top accuracy for each benchmark is shown in \textbf{bold}.
For each STR combination, we have run five trials with different initialization random seeds and have averaged their accuracies.}
	\label{sup:all_experiments}
\end{table*}

\begin{figure*}[!ht]
\centering
   \includegraphics[width=0.95\textwidth]{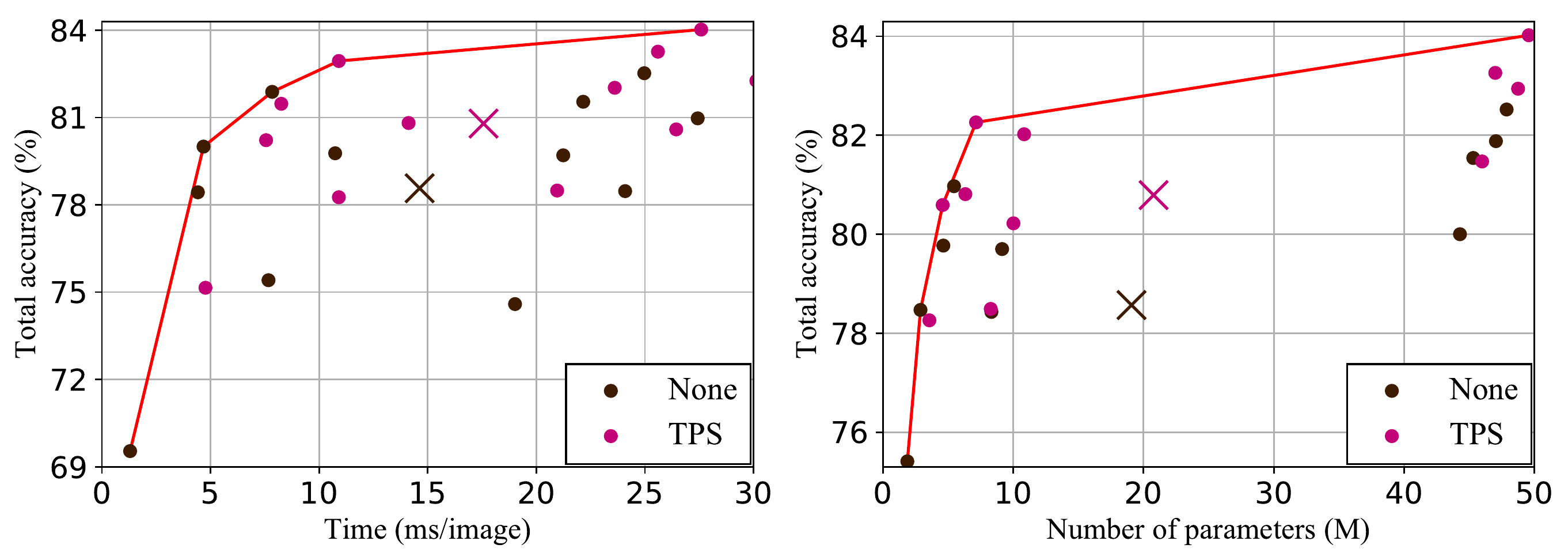}
    \caption{Color-coded version of Figure~4 in \S\ref{subsec:tradeoff}, according to the transformation stage.
    Each circle represents the performance for each different combination of STR modules, while the each cross represents the average performance among STR combinations without TPS (\textbf{\textcolor{black}{black}}) or with TPS (\textbf{\textcolor{magenta}{magenta}}).
    Choosing to add TPS or not does not seem to give a noticeable advantage in performance when looking at the performances of each STR combination.
    However, the average accuracy does increase when using TPS compared to when it is unused, at the cost of longer inference times and a slight increase in the number of parameters.
    }
  \label{sup:acc_time_parameter-trans}
\end{figure*}

\section{STR Framework - full experimental results} \label{sup:experiments}
\begin{figure*}[!ht]
\centering
   \includegraphics[width=0.95\textwidth]{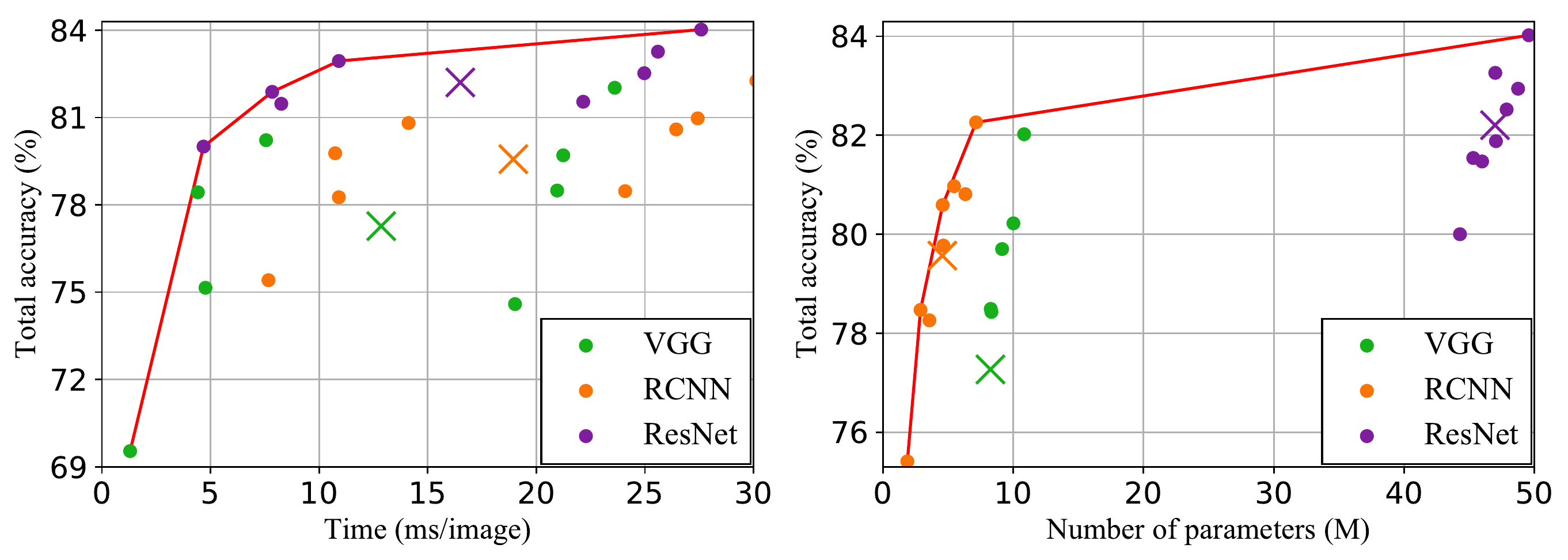}
    \caption{Color-coded version of Figure~4 in \S\ref{subsec:tradeoff}, according to the feature extraction stage.
    Each circle represents the performance for each different combination of STR modules, while the each cross represents the average performance among STR combinations using VGG (\textbf{\textcolor{green}{green}}), RCNN (\textbf{\textcolor{orange}{orange}}), or ResNet (\textbf{\textcolor{violet}{violet}}).
    VGG gives the lowest accuracy on average for the lowest amount of inference time required, while RCNN achieves higher accuracy over VGG by taking the longest time for inferencing and the lowest memory usage out of the three.
    ResNet, exhibits the highest accuracy at the cost of using significantly more memory than the other modules.
    In summary, if the system to be implemented is constrained by memory, RCNN offers the best trade-off, and if the system requires high accuracy, ResNet should be used.
    The time difference between the three modules are so small in practice that it should be considered only in the most extreme of circumstances.
    }
  \label{sup:acc_time_parameter-feat}
\end{figure*}

\begin{figure*}[!ht]
\centering
   \includegraphics[width=0.95\textwidth]{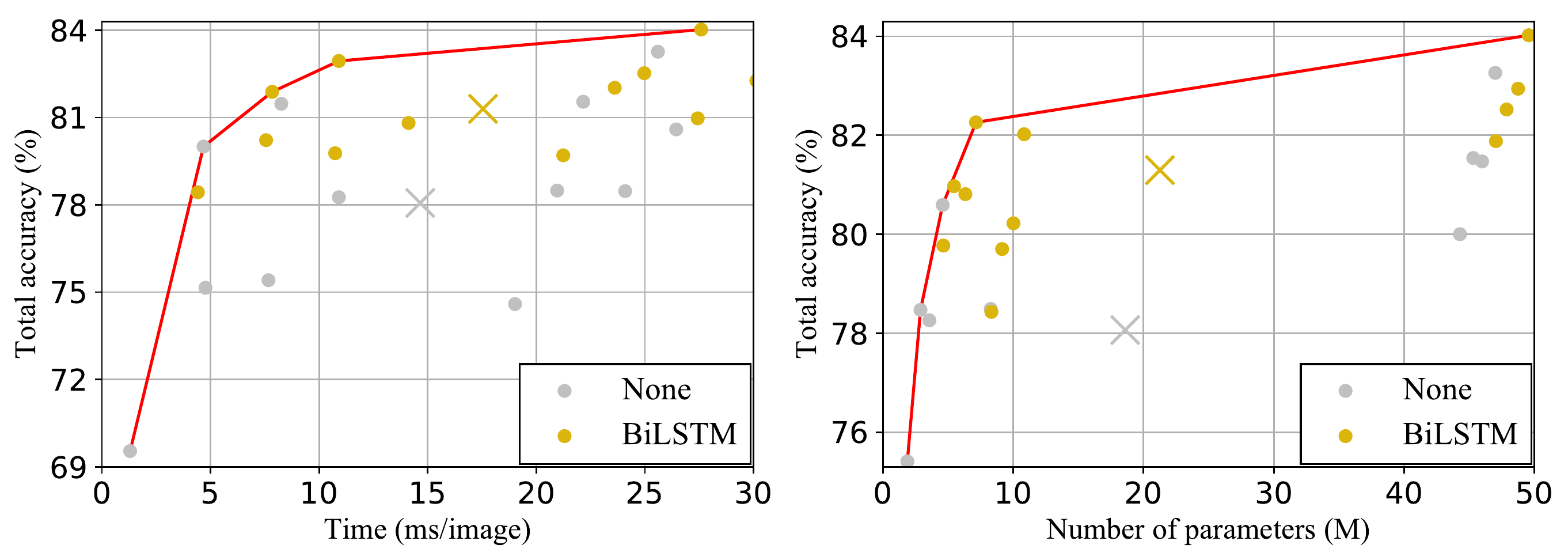}
    \caption{Color-coded version of Figure~4 in \S\ref{subsec:tradeoff}, according to the sequence modeling stage. 
    Each circle represents the performance for each different combination of STR modules, while the each cross represents the average performance among STR combinations without BiLSTM (\textbf{\textcolor{black!30}{gray}}) or with BiLSTM (\textbf{\textcolor{gold}{gold}}).
    Using BiLSTM seems to have a similar effect to using TPS, and vice versa, except BiLSTM gives a larger accuracy boost on average with similar inference time or parameter size concessions compared to TPS.
    }
  \label{sup:acc_time_parameter-seq}
\end{figure*}

\begin{figure*}[!ht]
\centering
   \includegraphics[width=0.95\textwidth]{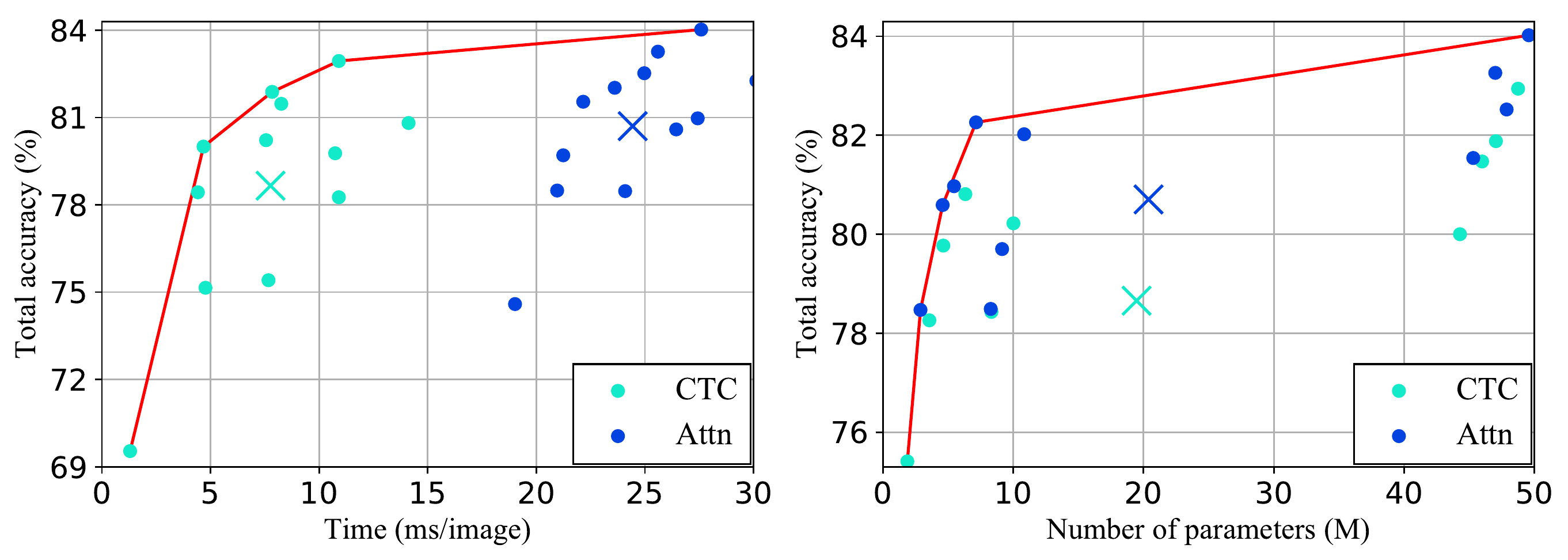}
    \caption{Color-coded version of Figure~4 in \S\ref{subsec:tradeoff}, according to prediction stage. 
    Each circle represents the performance for each different combination of STR modules, while the each cross represents the average performance among STR combinations with CTC (\textbf{\textcolor{cyan}{cyan}}) or with Attn (\textbf{\textcolor{blue}{blue}}).
    The choice between CTC and Attn gives the largest and clearest inference time increase for each percentage of accuracy gained.
    The same cannot be said about the increase in the number of parameters with respect to accuracy increase, as Attn gains about 2 percentage points with minimal memory usage increase.
    }
  \label{sup:acc_time_parameter-pred}
\end{figure*}

We report the full results of our experiments in Table~\ref{sup:all_experiments}.
FLOPS in Table~\ref{sup:all_experiments} is approximately calculated, the detail is in our GitHub issue\footnotemark.
In addition, Figure~\ref{sup:acc_time_parameter-trans}--\ref{sup:acc_time_parameter-pred} show two types of trade-off plots of 24 combinations in respect of accuracy versus time and accuracy versus memory.
In Figure~\ref{sup:acc_time_parameter-trans}--\ref{sup:acc_time_parameter-pred}, all the combination are color-coded in terms of each module, which helps to grasp the effectiveness of each module.

\footnotetext{\url{https://github.com/clovaai/deep-text-recognition-benchmark/issues/125}}

\clearpage 

\clearpage

\section{Additional Experiments} \label{sup:additional_experiments}
\subsection{Fine-tuning on real datasets}
We have fine-tuned our best model on 
the union of training sets IIIT, SVT, IC13, and IC15 (in-distribution), the held-out subsets of evaluation datasets of real scene text images.
Other evaluation datasets, IC03, SP, and CT (out-distribution), do not have held-out subset for training; SP and CT have not training sets and some training images of IC03 have been found in IC13 evaluation dataset, as mentioned in \S\ref{implementation}, thus it is not appropriate to fine-tuning on IC03 training set.

Our model has been fine-tuned for 10 epochs.
The table~\ref{sup:fine-tuning} shows the results.
By fine-tuning on the real data, the accuracy on in-distribution subset (the union of evaluation datasets IIIT, SVT, IC13, and IC15) and on all benchmark data have improved by 2.2 pp and 1.5 pp, respectively.
Meanwhile, the fine-tuned performance on the out-distribution subset (the union of evaluation datasets IC03, SP, and CT) has decreased by 1.3 pp.
We conclude that fine-tuning over real data is effective when the real-data is close to the test-time distribution. Otherwise, fine-tuning over real data may do more harm than good. 

\begin{table}[h]
    \centering
    \begin{adjustbox}{width=0.475\textwidth}
    \begin{tabular}{l|lll} 
    \hline
    \multirow{2}{*}{} & \multicolumn{3}{c}{Accuracy (\%)} \cr 
                 & in-distribution & out-distribution  & all \\
    \hline
    Original & 83.9 & 85.1 & 84.1 \\ 
    Fine-tuned  & 86.1(\textbf{+2.2}) & 83.8(\textit{-1.3}) & 85.6(\textbf{+1.5}) \\ 
    \hline
    \end{tabular}
    \end{adjustbox}
    \caption{Accuracy change with fine-tuning on real datasets.}
    \label{sup:fine-tuning}
\end{table}

\subsection{Accuracy with varying training dataset size} 
\begin{figure}[h]
\begin{center}
\includegraphics[width=0.95\linewidth]{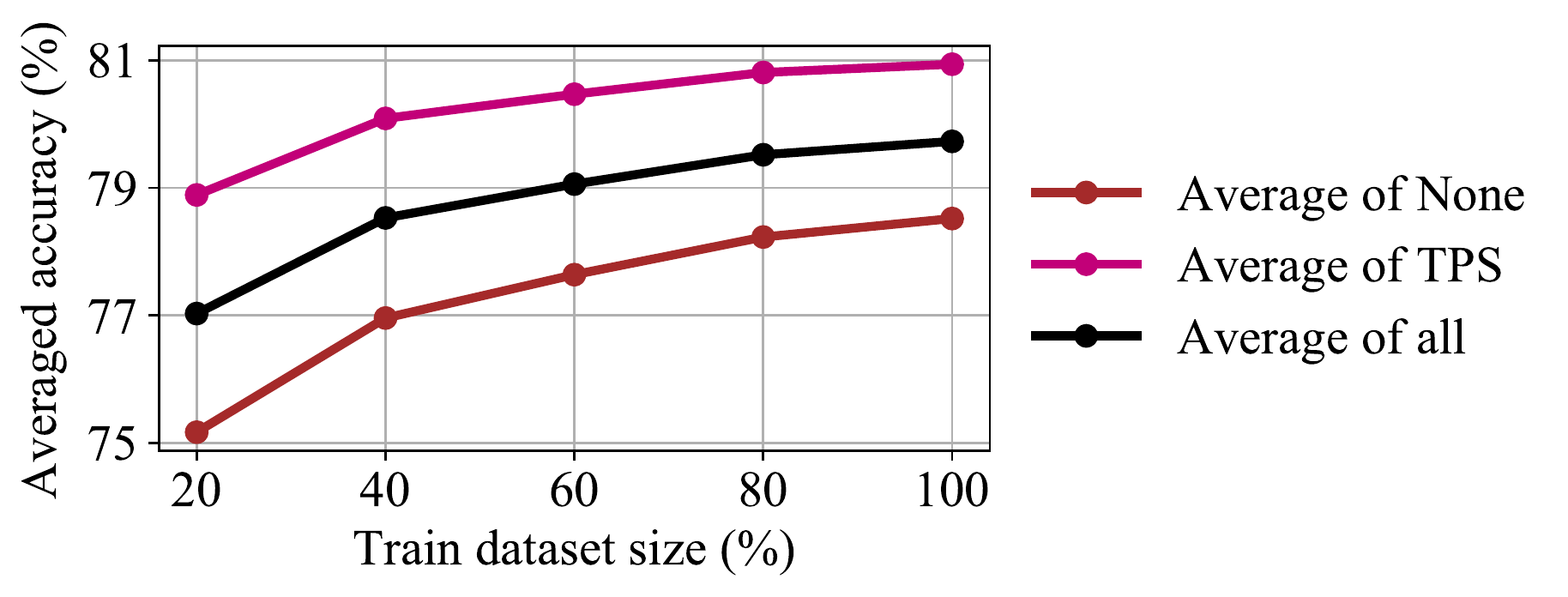}
\end{center}
\vspace{-6mm}
\caption{Averaged accuracy with varying training dataset size. Each plot represents the average performance among STR combinations without TPS (\textbf{\textcolor{brown}{brown}}), with TPS (\textbf{\textcolor{magenta}{magenta}}), or all models (\textbf{\textcolor{black}{black}})}
\label{sup:fig:DS-trans}
\end{figure}

\begin{figure}[h]
\begin{center}
\includegraphics[width=0.95\linewidth]{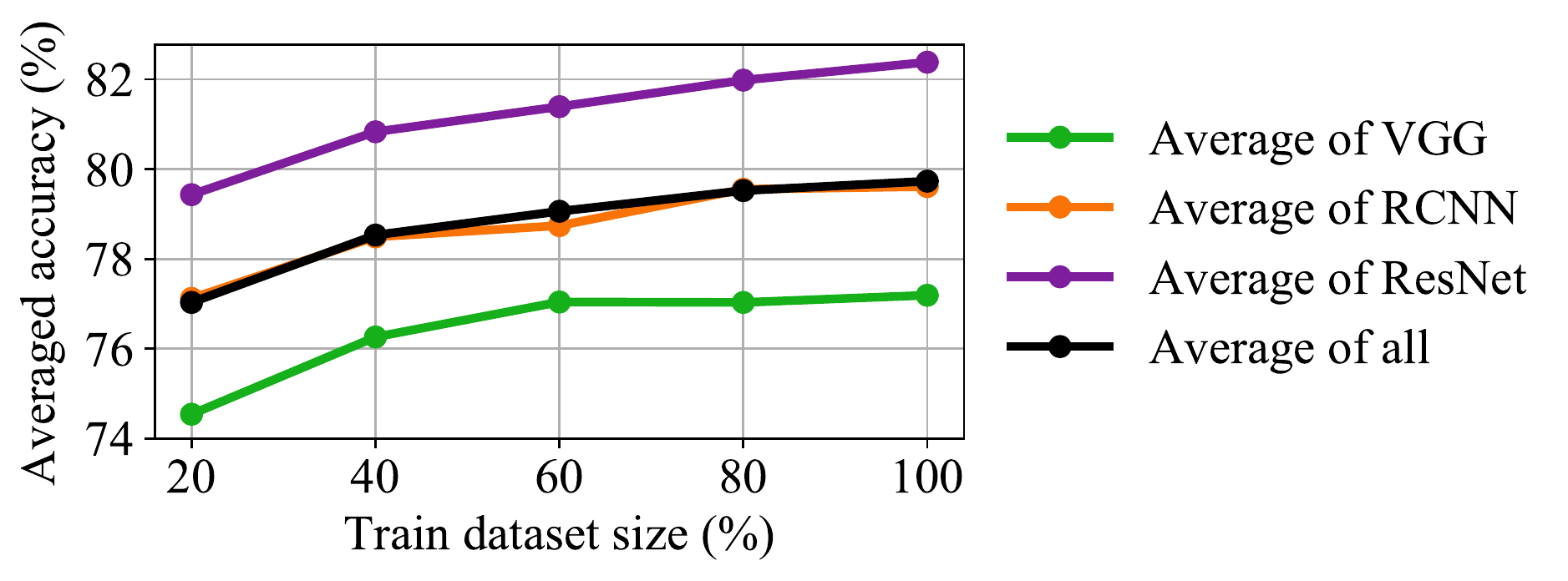}
\end{center}
\vspace{-6mm}
\caption{Averaged accuracy with varying training dataset size. Each plot represents the average performance among STR combinations using VGG (\textbf{\textcolor{green}{green}}), RCNN (\textbf{\textcolor{orange}{orange}}), ResNet (\textbf{\textcolor{violet}{violet}}), or all models (\textbf{\textcolor{black}{black}})}
\label{sup:fig:DS-feat}
\end{figure}

\begin{figure}[h]
\begin{center}
\includegraphics[width=0.95\linewidth]{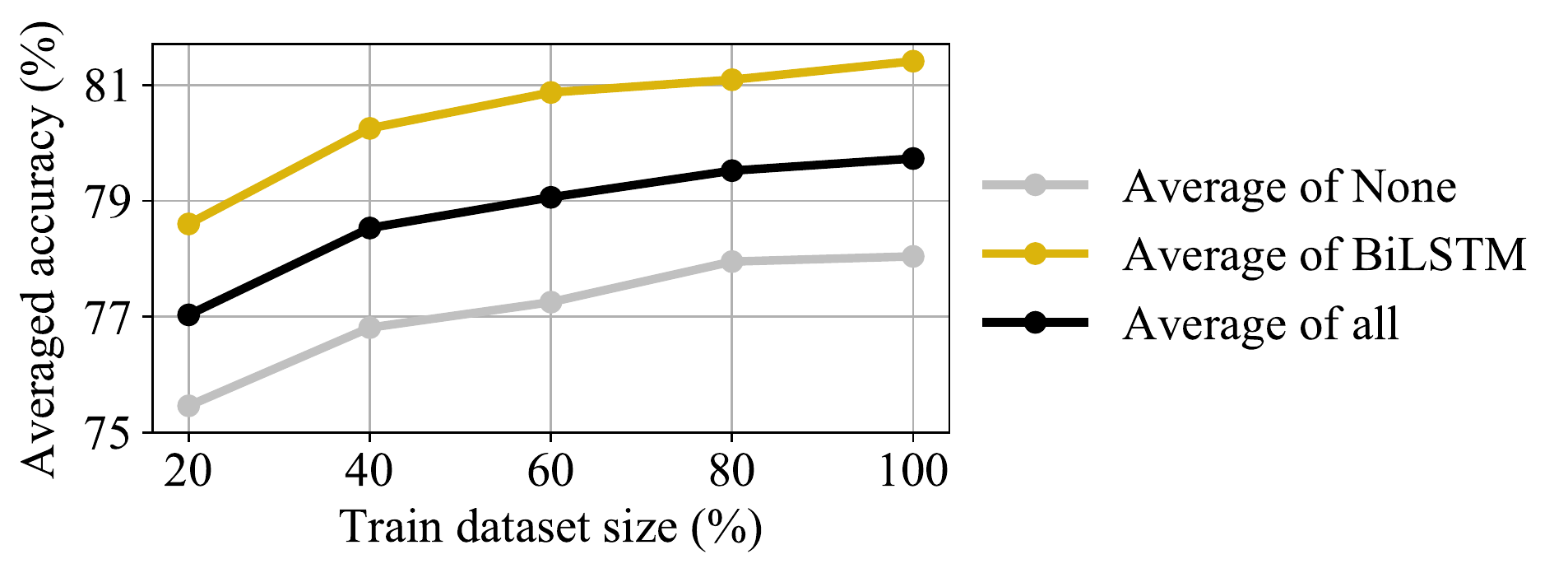}
\end{center}
\vspace{-6mm}
\caption{Averaged accuracy with varying training dataset size. Each plot represents the average performance among STR combinations without BiLSTM (\textbf{\textcolor{black!30}{gray}}), with BiLSTM (\textbf{\textcolor{gold}{gold}}), or all models (\textbf{\textcolor{black}{black}})}
\label{sup:fig:DS-seq}
\end{figure}

\begin{figure}[h]
\begin{center}
\includegraphics[width=0.95\linewidth]{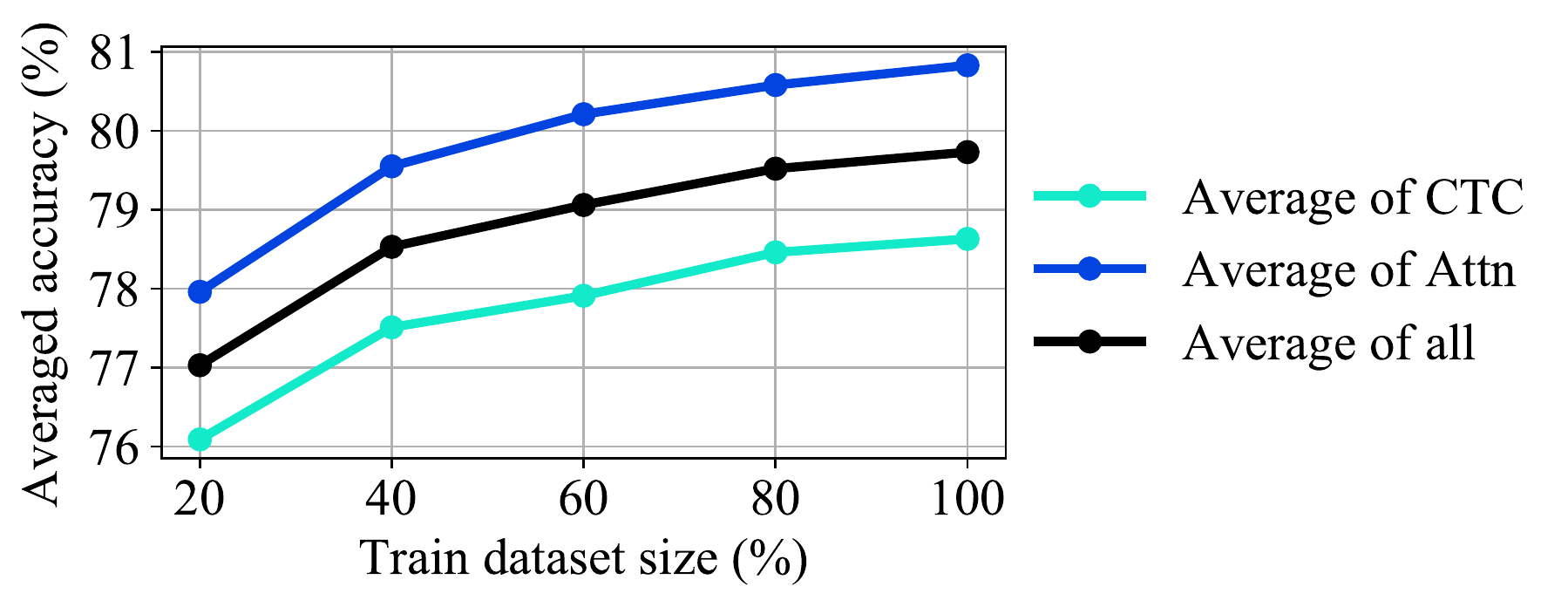}
\end{center}
\vspace{-6mm}
\caption{Averaged accuracy with varying training dataset size. Each plot represents the average performance among STR combinations with CTC (\textbf{\textcolor{cyan}{cyan}}), with Attn (\textbf{\textcolor{blue}{blue}}), or all models (\textbf{\textcolor{black}{black}})}
\label{sup:fig:DS-pred}
\end{figure}

We have evaluated the accuracy of all 24 STR models against varying training dataset size.
Training dataset consists of MJSynth 8.9~M and SynthText 5.5~M (14.4~M in total), same setting as in \S\ref{implementation}.
We report the full results of varying training dataset size in Table~\ref{sup:varying-DS-all-models}.
In addition, Figure~\ref{sup:fig:DS-trans}--\ref{sup:fig:DS-pred} show averaged accuracy plots. 
Each plot is color-coded in terms of each module, which helps to grasp the tendency of each module.

From the Table~\ref{sup:varying-DS-all-models} and the plot of average of all models in Figure~\ref{sup:fig:DS-trans}--\ref{sup:fig:DS-pred}, we observe that the average of all 24 models tends to have higher accuracy with more data.

In Figure~\ref{sup:fig:DS-trans}, we observe that the curves of without TPS do not get saturated at 100\% training data size; more training data are certainly likely to improve them. 
The curves of TPS show saturated performances at 80\% training data. 
We conjecture this is because TPS usually normalizes the input images and the last 20\% of training dataset would be normalized by TPS, rather than improve accuracy.
Thus other kinds of datasets, which will not simply be normalized by TPS trained with 80\% training dataset, would be needed to better accuracy.

In Figure~\ref{sup:fig:DS-feat}, we observe that the curves of ResNet do not get saturated at 100\% training data size. 
The averages of VGG and RCNN, on the other hand, show saturated performances at 60\% and 80\% training data, respectively.
We conjecture this is because VGG and RCNN have lower capacity than ResNet and they have already reached their performance limits at the current amount of training data. 

In Figure~\ref{sup:fig:DS-seq}, we observe that the curves of BiLSTM do not get saturated at 100\% training data size.
The curves of without BiLSTM show saturated performances at 80\% training data. 
We conjecture this is because using BiLSTM has higher capacity than without BiLSTM and thus using BiLSTM still has room for improving accuracy with more training data.

In Figure~\ref{sup:fig:DS-pred}, we observe that the curves of Attn do not get saturated at 100\% training data size.
The curves of CTC show saturated performances at 80\% training data. 
Again, we conjecture this is because using Attn has higher capacity than CTC and thus using Attn still has room for improving accuracy with more training data.


\begin{table}
	\centering\noindent
   	 \begin{adjustbox}{width=0.475\textwidth}
     \begin{threeparttable}
     \begin{tabular}{l|c|c|l|c|ccccc}
			\hline
  			\multirow{2}{*}{\#} & \multirow{2}{*}{Trans.} & \multirow{2}{*}{Feat.} & \multirow{2}{*}{Seq.} & \multirow{2}{*}{Pred.} &
			\multicolumn{5}{c}{Training dataset size (\%)} \cr
			&&&&& 20 & 40 & 60 & 80 & 100 \cr\hline
1 & \multirow{12}{*}{None} & \multirow{4}{*}{VGG} & \multirow{2}{*}{None} & CTC & 66.1 & 68.1 & 69.2 & 68.8 & 68.8\\
2 &  &  &  & Attn & 71.5 & 73.0 & 74.2 & 74.7 & 74.6\\\cline{4-10}
3\tnote{1} &  &  & \multirow{2}{*}{BiLSTM}& CTC & 75.8 & 77.6 & 77.7 & 77.9 & 78.6\\
4 &  &  &  & Attn & 75.6 & 77.9 & 79.3 & 79.3 & 79.7\\\cline{3-10}
5 &  & \multirow{4}{*}{RCNN} &  \multirow{2}{*}{None} & CTC & 69.7 & 71.2 & 72.0 & 75.5 & 74.7\\
6\tnote{2} &  &  &  & Attn & 76.2 & 77.5 & 77.3 & 78.1 & 78.2\\\cline{4-10}
7\tnote{3} &  &  & \multirow{2}{*}{BiLSTM} & CTC & 77.1 & 78.8 & 79.6 & 80.0 & 79.7\\
8 &  &  &  & Attn & 77.9 & 79.6 & 80.3 & 80.5 & 81.3\\\cline{3-10}
9\tnote{4} &  & \multirow{4}{*}{ResNet} &  \multirow{2}{*}{None} & CTC & 75.9 & 77.8 & 78.8 & 78.9 & 80.7\\
10 &  &  &  & Attn & 78.0 & 80.3 & 80.5 & 81.6 & 81.5\\\cline{4-10}
11 &  &  & \multirow{2}{*}{BiLSTM} & CTC & 78.9 & 80.7 & 80.8 & 81.3 & 81.7\\
12 &  &  &  & Attn & 79.2 & 81.0 & 81.9 & 82.3 & 82.6\\
\hline
13 & \multirow{12}{*}{TPS} & \multirow{4}{*}{VGG} &  \multirow{2}{*}{None} & CTC & 73.8 & 74.9 & 75.4 & 75.5 & 75.2\\
14 &  &  &  & Attn & 75.9 & 78.3 & 78.8 & 78.5 & 78.7\\\cline{4-10}
15 &  &  & \multirow{2}{*}{BiLSTM} & CTC & 77.9 & 79.2 & 79.9 & 79.5 & 80.1\\
16\tnote{5} &  &  &  & Attn & 79.6 & 81.1 & 81.7 & 82.0 & 81.9\\\cline{3-10}
17 &  & \multirow{4}{*}{RCNN} &  \multirow{2}{*}{None} & CTC & 77.8 & 78.5 & 76.8 & 78.6 & 78.1\\
18 &  &  &  & Attn & 79.2 & 79.8 & 80.5 & 80.4 & 80.7\\\cline{4-10}
19 &  &  & \multirow{2}{*}{BiLSTM} & CTC & 78.7 & 80.7 & 81.2 & 80.8 & 80.9\\
20 &  &  &  & Attn & 80.4 & 81.8 & 82.2 & 82.5 & 83.1\\\cline{3-10}
21 &  & \multirow{4}{*}{ResNet} &  \multirow{2}{*}{None} & CTC & 80.7 & 80.7 & 80.8 & 81.7 & 81.9\\
22 &  &  &  & Attn & 80.7 & 81.6 & 82.6 & 83.0 & 83.3\\\cline{4-10}
23\tnote{6} &  &  & \multirow{2}{*}{BiLSTM} & CTC & 80.7 & 81.8 & 82.6 & 83.0 & 83.2\\
24\tnote{7} &  &  &  & Attn & 81.3 & 82.7 & 83.2 & 84.0 & 84.1\\
\hline
		\end{tabular}
		\begin{tablenotes}
            \item[]$^1$ CRNN. $^2$ R2AM. $^3$ GRCNN. $^4$ Rosetta. $^5$ RARE. $^6$ STAR-Net. $^7$ our best model.
  		\end{tablenotes}
        \end{threeparttable}
        \end{adjustbox}
     \captionof{table}{The full experimental results of varying training dataset size for all 24 STR combinations. 
     Each value represent Total accuracy (\%), as mentioned in \S\ref{implementation}.
     Note that, for each STR combination, we have run only one trial and thus the result could be slightly different from the Table~\ref{sup:all_experiments}.}
	\label{sup:varying-DS-all-models}
\end{table}

\vspace{-3mm}
\begin{figure*}[htbp]
\begin{center}
\includegraphics[width=0.95\textwidth]{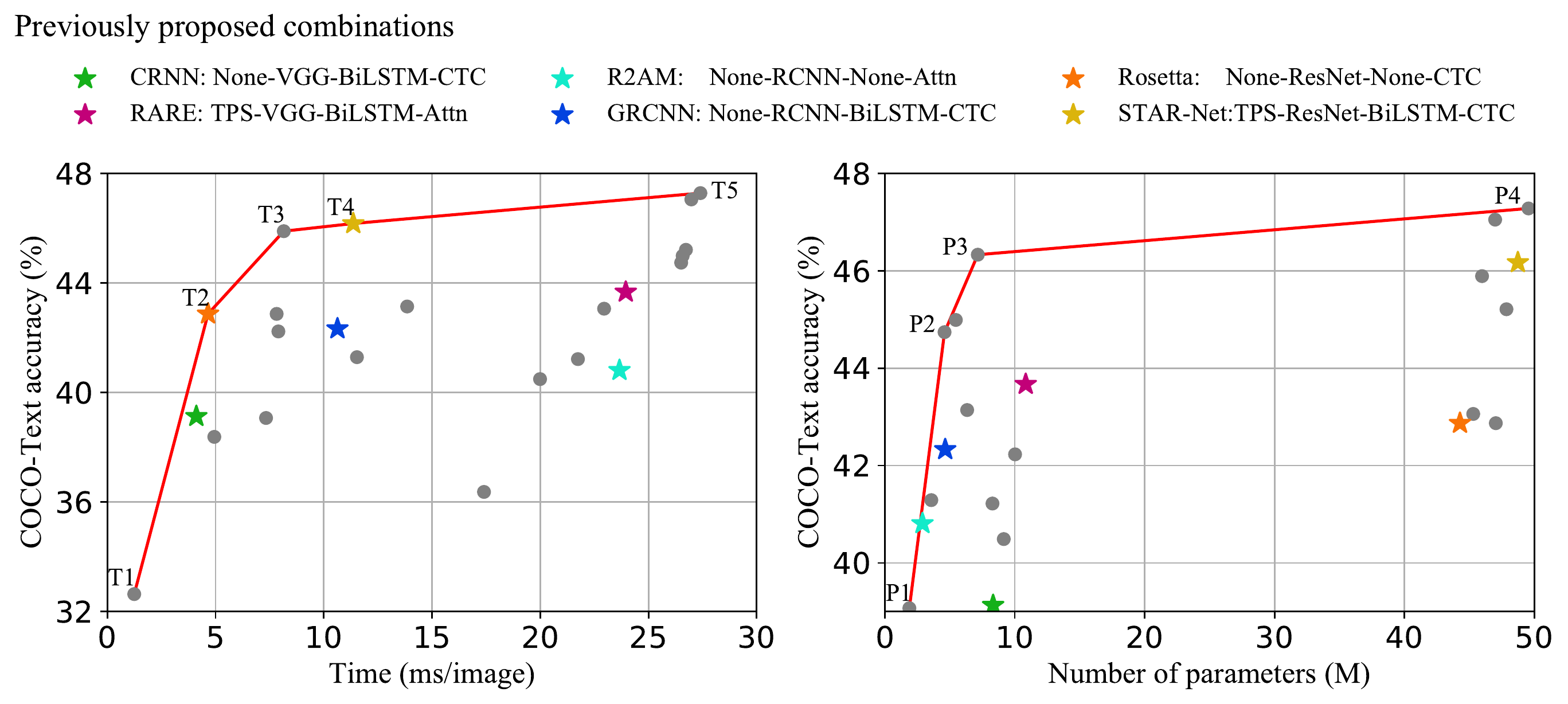}
\end{center}
\caption{COCO-Text accuracy version of Figure~4 in \S\ref{subsec:tradeoff}.}
\label{sup:fig:COCO}
\end{figure*}

\newpage

\subsection{Evaluation on COCO-Text dataset}
We have evaluated the models on COCO-Text dataset~\cite{COCO}, another good benchmark derived from MS COCO containing complex and low-resolution scene images. COCO-Text contains many special characters, heavy noises, and occlusions; it is generally considered more challenging than the seven benchmarks considered so far.
Figure~\ref{sup:fig:COCO} shows the accuracy-time and accuracy-space trade-off plots for 24 STR methods on COCO-Text.
Except that the overall accuracy is lower, the relative orders amongst methods are largely preserved compared to Figure~4. 
Fine-tuning models with COCO-Text training set has improved the averaged accuracy (24 models) from 42.4\% to 58.2\%, a relatively big jump that is attributable to the unusual data distribution for COCO-Text.
Evaluation and analysis over COCO-Text are beneficial, especially to address remaining corner cases for STR.

\end{document}